%% file: top_full.tex
\documentclass[final]{cvpr}

\usepackage{times}
\usepackage{epsfig}
\usepackage{float}
\usepackage{graphicx}
\usepackage{algorithm}
\usepackage{algorithmicx}
\usepackage{algpseudocode}
\usepackage{xcolor}
\usepackage{amsmath}
\usepackage{amssymb}
\usepackage{booktabs}
\usepackage{subcaption}
\usepackage{balance}
\usepackage{verbatim}

\usepackage[pagebackref=true,breaklinks=true,colorlinks,bookmarks=true]{hyperref}

\def\cvprPaperID{875} %
\def\confYear{CVPR 2021}

\input{shortcuts.tex}

\begin{document}

\title{Stable View Synthesis}
\author{Gernot Riegler \quad\quad Vladlen Koltun \\Intel Labs}

\makeatletter
\g@addto@macro\@maketitle{
  \begin{figure}[H]
  \center
  \setlength{\linewidth}{\textwidth}
  \setlength{\hsize}{\textwidth}
  \vspace{-2.25em}
  \includegraphics[width=\linewidth]{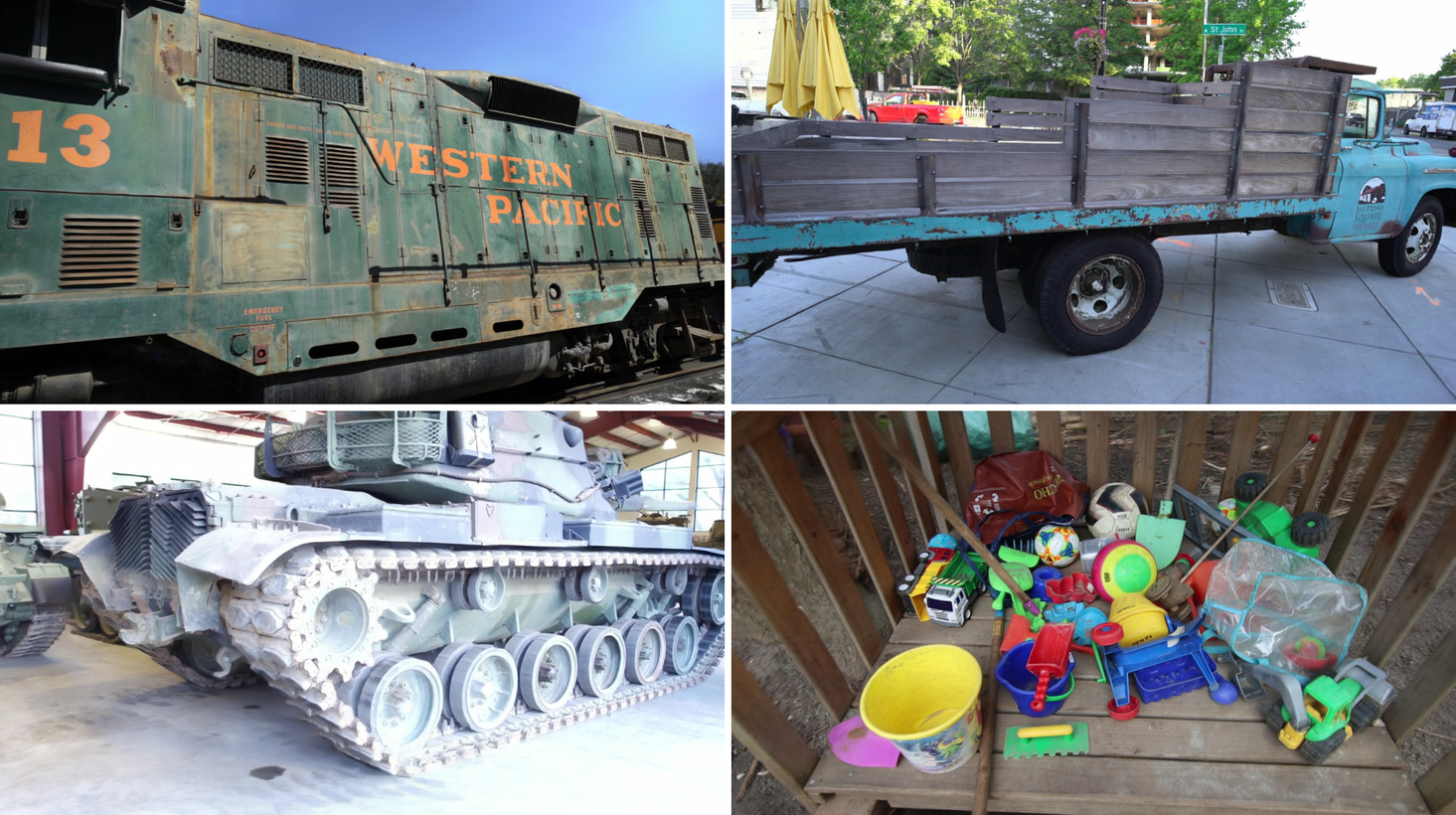}
  \vspace{-0.75em}
  \caption{
    Stable View Synthesis synthesizes spatially and temporally coherent photorealistic views of complex real-world scenes.
    Top and left: new views of scenes from the Tanks and Temples dataset~\cite{Knapitsch2017Tanks}. Bottom right: a new view of a scene from the FVS dataset~\cite{Riegler2020Free}.
  }
  \label{fig:teaser}
  \end{figure}
}
\makeatother

\maketitle

\input{top_sec_00-abstract.tex}

\cleardoublepage

\input{top_sec_01-intro.tex}

\input{top_sec_02-related.tex}

\input{top_sec_03-method.tex}
\input{top_sec_04-eval.tex}

\input{top_sec_05-conclusion.tex}

\balance

{\small
\bibliographystyle{ieee_fullname}
\bibliography{egbib_shortstr,egbib}
}

\newpage
\section{Appendix}
\newcommand{\mysuppsection}{\subsection}

\input{top_sec_06-appendix.tex}

\end{document}

%% file: shortcuts.tex
\definecolor{targetcolor}{RGB}{249, 65, 68}
\definecolor{color}{RGB}{243, 114, 44}
\definecolor{color}{RGB}{248, 150, 30}
\definecolor{color}{RGB}{249, 199, 79}
\definecolor{color}{RGB}{144, 190, 109}
\definecolor{enccolor}{RGB}{67, 170, 139}
\definecolor{pointcolor}{RGB}{87, 117, 144}

\newcommand{\myunaryless}{{<}}

\DeclareMathOperator{\mean}{mean}
\DeclareMathOperator{\aggr}{\nu}

\newcommand{\myreals}{\mathbb{R}}

\newcommand{\myvec}[1]{\mathbf{#1}}
\newcommand{\mymat}[1]{\mathbf{#1}}
\newcommand{\myten}[1]{\mathcal{#1}}

\newcommand{\myimg}{\myten{I}}
\newcommand{\mynimgs}{N}
\newcommand{\myimgn}{n}
\newcommand{\mykimgs}{K}
\newcommand{\myimgk}{k}
\newcommand{\mymimgs}{M}
\newcommand{\myimgm}{m}
\newcommand{\mytgtt}{t}

\newcommand{\mynimgsset}[1]{\{#1_\myimgn\}_{\myimgn=1}^{\mynimgs}}
\newcommand{\mykimgsset}[1]{\{#1_\myimgk\}_{\myimgk=1}^{\mykimgs}}
\newcommand{\myimgset}{\mynimgsset{\myimg}}

\newcommand{\myenc}{\myten{F}}
\newcommand{\mytgt}{\myten{G}}
\newcommand{\myout}{\myten{O}}
\newcommand{\mydepthmap}{\myten{D}}

\newcommand{\mycammat}{\mymat{K}}
\newcommand{\myrotmat}{\mymat{R}}
\newcommand{\mytransvec}{\myvec{t}}

\newcommand{\mysurface}{\Gamma}

\newcommand{\mysurfpoint}{\myvec{x}}
\newcommand{\myencvec}{\myvec{f}}
\newcommand{\mytgtvec}{\myvec{g}}
\newcommand{\myviewdir}{\myvec{v}}
\newcommand{\mytgtviewdir}{\myvec{u}}

\newcommand{\mysurfpointset}{\{\mysurfpoint_{h,w}\}_{h,w=1,1}^{H \times W}}
\newcommand{\mysurfpoints}{\myten{X}}

\newcommand{\myencnet}{\phi_\text{enc}}
\newcommand{\myaggrfcn}{\phi_\text{aggr}}
\newcommand{\myrendernet}{\phi_\text{render}}

\newcommand{\mynetweights}{\theta}
\newcommand{\myencweights}{\mynetweights_\text{enc}}
\newcommand{\myaggrweights}{\mynetweights_\text{aggr}}
\newcommand{\myrenderweights}{\mynetweights_\text{render}}
\newcommand{\myimgweights}{\mynetweights_\text{imgs}}

\newcommand{\myloss}{\mathbb{L}}

\newcommand{\boldparagraph}[1]{\vspace{1em}\noindent{\bf #1:} }

%% file: top_sec_00-abstract.tex
\begin{abstract}
We present Stable View Synthesis (SVS). Given a set of source images depicting a scene from freely distributed viewpoints, SVS synthesizes new views of the scene. The method operates on a geometric scaffold computed via structure-from-motion and multi-view stereo. Each point on this 3D scaffold is associated with view rays and corresponding feature vectors that encode the appearance of this point in the input images. The core of SVS is view-dependent on-surface feature aggregation, in which directional feature vectors at each 3D point are processed to produce a new feature vector for a ray that maps this point into the new target view. The target view is then rendered by a convolutional network from a tensor of features synthesized in this way for all pixels. The method is composed of differentiable modules and is trained end-to-end. It supports spatially-varying view-dependent importance weighting and feature transformation of source images at each point; spatial and temporal stability due to the smooth dependence of on-surface feature aggregation on the target view; and synthesis of view-dependent effects such as specular reflection. Experimental results demonstrate that SVS outperforms state-of-the-art view synthesis methods both quantitatively and qualitatively on three diverse real-world datasets, achieving unprecedented levels of realism in free-viewpoint video of challenging large-scale scenes.
Code is available at \url{https://github.com/intel-isl/StableViewSynthesis}
\end{abstract}

%% file: top_sec_01-intro.tex
\section{Introduction}

Photorealistic view synthesis can allow us to explore magnificent sites in faraway lands without leaving the comfort of our homes.
This requires advancing the technology towards two key goals. First, the synthesized images should be photorealistic: indistinguishable from reality. Second, the user should be free to move through the scene, as in the real world, exploring it from any physically realizable viewpoint.

In this paper, we present a new method for photorealistic view synthesis that brings these two goals closer. Our input is a set of images that can be taken for example from a handheld video of the scene. From these images, we construct a 3D geometric scaffold via off-the-shelf structure-from-motion, multi-view stereo, and meshing. Input images are encoded by a convolutional network and the resulting deep features are mapped onto the geometric scaffold. As a result, for any point on the scaffold, we can obtain a collection of view rays with associated feature vectors, which correspond to input images that see this point.

The core of our method is an approach to synthesizing arbitrary new views given this representation of the scene. Each pixel in the new view is mapped onto the geometric scaffold to obtain the set of input rays with associated feature vectors, and an output ray towards the new view. The feature vectors from the input rays are then aggregated, taking the geometry of the input and output rays into account, by a differentiable module that produces a feature vector for the output ray.
Together, the feature vectors synthesized for all pixels form a feature tensor. The new image is rendered from this feature tensor by a convolutional network.

All steps of the method are differentiable and the complete pipeline can be trained end-to-end to maximize photorealism. All steps can be implemented efficiently, leveraging parallelism across pixels.
Crucially, the computation of a feature vector for a new output ray does not require any heuristic selection of input rays. The computation aggregates information from all input rays in a differentiable module that is informed by the spatial layout of the rays and is optimized end-to-end. This supports temporal stability for smoothly moving viewpoints.

We evaluate the presented method on three diverse datasets of real scenes and objects: Tanks and Temples~\cite{Knapitsch2017Tanks}, FVS~\cite{Riegler2020Free}, and DTU~\cite{Aanaes2016DTU}. Tanks and Temples and FVS provide handheld video sequences of large real-world scenes; the objective is to use these video sequences as input to enable photorealistic rendering of the scenes from new views. DTU provides regularly-spaced outside-in images of challenging real objects. On all three datasets, SVS convincingly outperforms the state of the art. On Tanks and Temples, our method reduces the LPIPS error for new views by up to 10 absolute percentage points (a reduction of roughly 30\% on average) relative to the prior state of the art,
while also improving PSNR and SSIM. On the FVS dataset, our method likewise outperforms the state of the art on all metrics, reducing LPIPS by 7 absolute percentage points on average relative to the best prior method.
On DTU, we set the new state of the art for novel view synthesis, attaining an average LPIPS error of 4.5\% over the test scenes in extrapolation mode and 1.6\% for view interpolation. A number of our synthesized images for new views in Tanks and Temples and FVS scenes are shown in Figure~\ref{fig:teaser}, and video sequences are provided in the supplementary video.

%% file: top_sec_02-related.tex
\section{Related Work}

\begin{figure*}[t]
    \center
    \begin{subfigure}[b]{0.24\textwidth}
        \includegraphics[width=\textwidth]{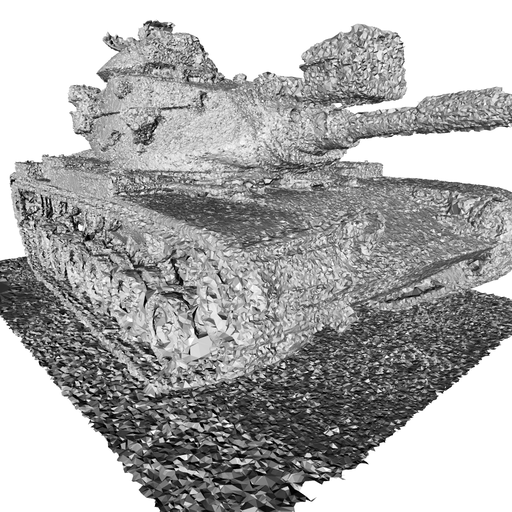}
        \caption{Geometric scaffold}
        \label{fig:arch_scaffold}
    \end{subfigure}
    \begin{subfigure}[b]{0.23\textwidth}
        \includegraphics[width=\textwidth]{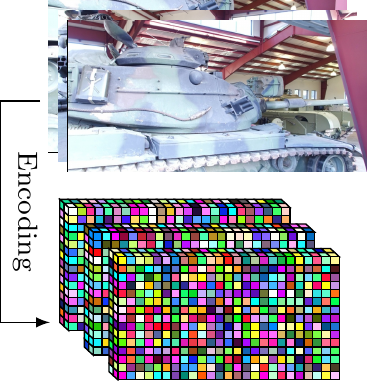}
        \caption{Encoding source images}
        \label{fig:arch_encode}
    \end{subfigure}
    \begin{subfigure}[b]{0.26\textwidth}
        \includegraphics[width=\textwidth]{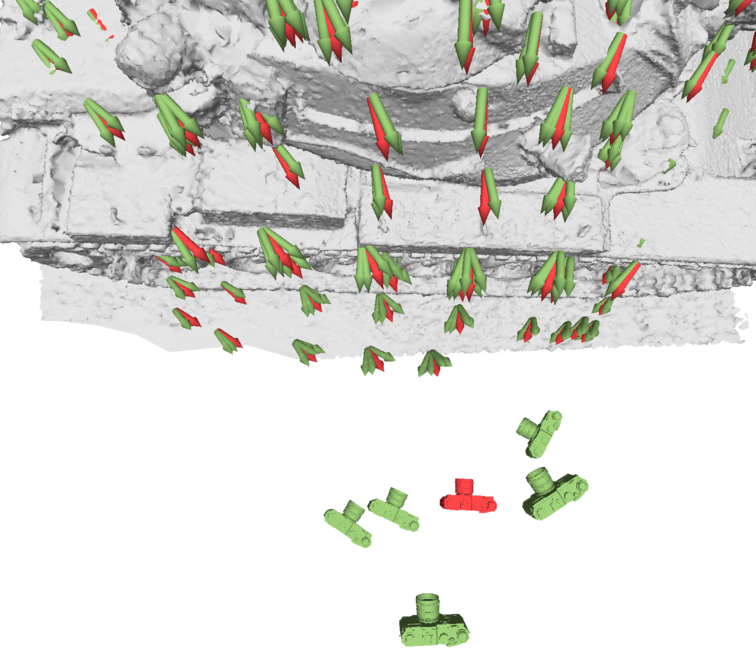}
        \caption{On-surface aggregation}
        \label{fig:arch_aggr}
    \end{subfigure}
    \begin{subfigure}[b]{0.23\textwidth}
        \includegraphics[width=\textwidth]{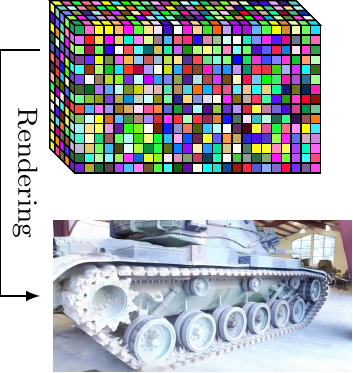}
        \caption{Rendering in the target view}
        \label{fig:arch_decode}
    \end{subfigure}
    \caption{
        \textbf{Overview of Stable View Synthesis.}
        (a) A geometric scaffold of the scene is constructed using structure-from-motion, multiple-view stereo, and meshing.
        (b) All source images are encoded into feature tensors via a convolutional network.
        (c) Given a new target view (red camera), feature vectors from the source images (green cameras) are aggregated on the geometric scaffold.
        Red arrows map 3D points to the target view, green arrows map the same points to the source views.
        (d) The output image in the target view is rendered from a tensor of synthesized feature vectors by a convolutional network.
    }
    \label{fig:arch}
    \vspace{-1em}
\end{figure*}

Image-based rendering has a long history in computer vision and graphics. Shum and Kang~\cite{Shum2000review} provide a review of early approaches and foundational work. More recent highlights include the work of Wood et al.~\cite{Wood2000Surface}, Buehler et al.~\cite{Buehler2001Unstructured}, Davis et al.~\cite{Davis2012Unstructured}, Chaurasia et al.~\cite{Chaurasia2013depthSynthesis}, Kopf et al.~\cite{Kopf2014First}, Hedman et al.~\cite{Hedman2016Scalable}, and Penner and Zhang~\cite{Penner2017Soft3D}.

More recently, deep learning techniques have enabled a new level of flexibility and realism.
Given a geometric reconstruction of the scene, Hedman et al.~\cite{Hedman2018Deep} map image mosaics to the target view and refine them via a blending network.
Thies et al.~\cite{Thies2020Ignor} learn image-dependent effects via a convolutional network.
Choi et al.~\cite{Choi2019Extreme} warp volumetric information from the source images to the target view.
Riegler and Koltun~\cite{Riegler2020Free} warp features from a heuristically selected set of source images into the target view and blend them using a recurrent convolutional network.
Other approaches directly learn features for each 3D point~\cite{Aliev2020Neural,Dai2020Neural} or vertex~\cite{Thies2019Deferred} of a geometric reconstruction.

Our method is most closely related to the Free View Synthesis approach of Riegler and Koltun~\cite{Riegler2020Free}, in that both methods operate on a geometric scaffold obtained via SfM, MVS, and meshing, and both methods utilize encoder and decoder networks to encode input images into feature tensors and render the new view from a new feature tensor, respectively. However, the methods differ crucially at their core: the synthesis of the feature tensor for the new view. The FVS pipeline heuristically selects a set of relevant source images for a given target view, warps the feature tensors from these input views into the target camera frame, and blends these warped feature tensors via a recurrent convolutional network. The heuristic selection of relevant input views leads to temporal instability when the set of selected views changes and causes drastic visual artifacts when the selected views do not contain all the information needed to cover some part of the output image. Furthermore, the sequential ordering of the input feature tensors processed by the recurrent network is artificial and can lead to instability when it changes. In contrast, SVS synthesizes feature vectors for the new view on the 3D surface itself, taking all input images into account as needed, and using set operators rather than sequence models to avoid arbitrary ordering. There is no heuristic selection of relevant images, no temporal instability due to changes in this set, no drastic artifacts due to the heuristic omission of relevant information, and no instability due to shifts in sequential processing. All processing takes all available information into account as needed, via permutation-invariant set operators, in a pipeline that is composed entirely of differentiable modules that are trainable end-to-end.

Several methods incorporate concepts similar to plane-sweep volumes~\cite{Collins1996Space} into the network architecture to synthesize novel views.
Flynn et al.~\cite{Flynn2016DeepStereo} utilize this concept to interpolate between views.
Kalantari et al.~\cite{Kalantari2016Learning} use this idea for a light-field setup with a fixed number of cameras.
Additional directional lighting extensions to these architectures enable synthesis of complex appearance effects~\cite{Bi2020Deep3DCapture,Xu2019DeepViewSynthesis}.

Multi-plane images~(MPIs)~\cite{Zitnick2004High} are also often used in conjunction with deep networks~\cite{Zhou2018StereoMagnification}. Here the image is represented by color$+\alpha$ planes at different depths and novel views can be rendered back-to-front.
Srinivasan et al.~\cite{Srinivasan2019PushingBoundaries} show that a limiting factor in MPIs is the depth resolution and propose a randomized-resolution training procedure.
This work is extended by Mildenhall et al.~\cite{Mildenhall2019Local} who use multiple local MPIs and practical user guidance.
Flynn et al.~\cite{Flynn2019Deepview} train a network to predict high-quality MPIs via learned gradient descent.
Li et al.~\cite{Li2020Crowdsampling} extend this line of work to image sets with strong appearance variation.

Another class of methods utilizes volumetric representations.
Sitzmann et al.~\cite{Sitzmann2019Deepvoxels} lift 2D image features to a common 3D volume.
The features are synthesized via a scene-dependent rendering network.
To overcome the memory requirements of voxel-based representations, Lombardi et al.~\cite{Lombardi2019NeuralVolumes} learn a dynamic irregular grid structure.
In Scene Representation Networks~\cite{Sitzmann2019Scene}, the volume is represented as an MLP and images are rendered via differentiable ray marching.
Niemeyer et al.~\cite{Niemeyer2020Differentiable} build upon an implicit occupancy representation that can be trained by posed images via implicit differentiation.
Neural Radiance Fields~\cite{Mildenhall2020Nerf} produce impressive results by training an MLP that maps 3D rays to occupancy and color.
Images are synthesized from this representation via volume rendering.
This methodology has been extended to unbounded outdoor scenes~\cite{Zhang2020Nerfpp} and crowdsourced image collections~\cite{Martin2020NeRFW}.

%% file: top_sec_03-method.tex
\section{Overview}

A visual overview of SVS is provided in Figure~\ref{fig:arch}.
Our input is a set of source images $\myimgset$, which are used to erect a geometric scaffold $\mysurface$ and are the basis for the on-surface feature representation.
Given a new viewpoint $(\myrotmat_\mytgtt, \mytransvec_\mytgtt)$ and camera intrinsics $\mycammat_\mytgtt$, our goal is to synthesize an image $\myout$ that depicts the scene in this new view.

\boldparagraph{Preprocessing}
Our method leverages a 3D geometric scaffold.
To construct this scaffold, we use standard structure-from-motion, multi-view stereo, and surface reconstruction~\cite{Schoenberger2016SfM,Schoenberger2016Pixelwise}.
We first run structure-from-motion~\cite{Schoenberger2016SfM} to get camera intrinsics $\mynimgsset{\mycammat}$ and camera poses as rotation matrices $\mynimgsset{\myrotmat}$ and translation vectors $\mynimgsset{\mytransvec}$.
In the rest of the paper, we use $\myimgset$
to denote the rectified images after structure-from-motion.
We then run multi-view stereo on the posed images, obtain per-image depthmaps, and fuse these into a point cloud.
Delaunay-based 3D surface reconstruction is applied to this point cloud to get a 3D surface mesh $\mysurface$.
We use COLMAP~\cite{Schoenberger2016SfM,Schoenberger2016Pixelwise} for preprocessing in all experiments, but our method can utilize other SfM and MVS pipelines.

In addition, each image $\myimg_\myimgn$ is encoded by a convolutional network to obtain a feature tensor $\myenc_\myimgn$, which provides a feature vector for each pixel in $\myimg_\myimgn$.

\begin{figure}[t]
    \center
    \includegraphics[width=0.6\linewidth]{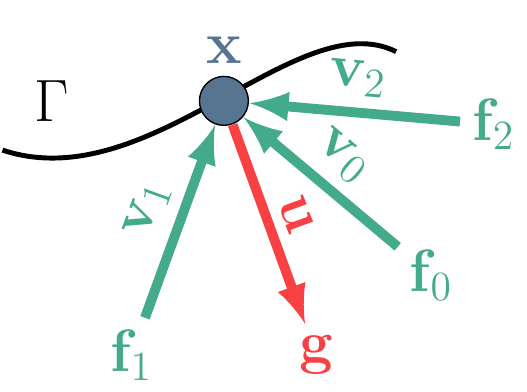}
    \vspace{-0.5em}
    \caption{
        \textbf{On-surface aggregation.}
        A 3D point $\mysurfpoint$ on the geometric scaffold $\mysurface$ is seen in a set of source images. Each such image contributes a feature vector $\myencvec_\myimgk$ along a ray $\myviewdir_\myimgk$ (green). On-surface aggregation uses a differentiable set network to process this data and produces a feature vector $\mytgtvec$ for the target ray $\mytgtviewdir$ (red).
    }
    \label{fig:aggr3d}
    \vspace{-1.5em}
\end{figure}

\boldparagraph{View synthesis}
To synthesize the new view $\myout$,
we backproject pixels in $\myout$ onto the scaffold $\mysurface$. For each point $\mysurfpoint\in \mysurface$ obtained in this way, we query the set of input images in which $\mysurfpoint$ is visible. For each such image $\myimg_\myimgk$, we obtain a feature vector $\myencvec_\myimgk$ along the corresponding ray $\myviewdir_\myimgk$ to $\mysurfpoint$.
See Figure~\ref{fig:aggr3d} for an illustration.
The set $\{(\myviewdir_\myimgk, \myencvec_\myimgk)\}_\myimgk$ of view rays with corresponding feature vectors is then processed by a differentiable set network that is conditioned on the output view direction $\mytgtviewdir$. This network produces a new feature vector $\mytgtvec$.
Feature vectors $\mytgtvec$ are obtained in this way for all pixels in $\myout$. The resulting feature tensor $\mytgt$ is decoded by a convolutional network to produce the output image.

Note that SVS differs from works that use neural point features~\cite{Aliev2020Neural,Dai2020Neural} or neural mesh textures~\cite{Thies2019Deferred}, which fit feature vectors from scratch (initialized with random noise) per scene on a point cloud or mesh.
SVS also differs from methods that project full (encoded) source images to the target view~\cite{Hedman2018Deep,Riegler2020Free}; in SVS, each 3D point independently aggregates features from a different set of source images.

\section{Feature Processing and Aggregation}

\vspace{-1em}
\boldparagraph{Image encoding}
Each source image $\myimg_\myimgn$ is encoded into a feature tensor by a convolutional network based on the U-Net architecture~\cite{Ronneberger2015UNet}. This network is denoted by $\myencnet$.
The encoder part of $\myencnet$ consists of an ImageNet-pretrained ResNet18~\cite{He2016Deep}, where we freeze the BatchNorm parameters.
In the decoder part of $\myencnet$, each stage upsamples the feature map using nearest-neighbor interpolation, concatenates it with the corresponding feature map (of the same resolution) from the encoder, and applies convolution and activation layers.
We denote the feature tensor produced by this network by $\myenc_\myimgn = \myencnet(\myimg_\myimgn)$.

\boldparagraph{On-surface aggregation}
The core of our method is the computation of a target feature vector $\mytgtvec(\mysurfpoint, \mytgtviewdir)$ for each point $\mysurfpoint \in \mysurface \subset \myreals^3$ on the 3D geometric scaffold.
This feature vector is computed as a function of the viewing direction $\mytgtviewdir$ from the target camera center
to the surface point $\mysurfpoint$, and tuples $\{(\myviewdir_\myimgk, \myencvec_\myimgk(\mysurfpoint))\}_{\myimgk=1}^\mykimgs$. Here $\{\myencvec_\myimgk(\mysurfpoint)\}_{\myimgk=1}^\mykimgs$ are source image features that correspond to $\mysurfpoint$ in the image encodings $\{\myenc_\myimgk\}_{\myimgk=1}^\mykimgs$ in which $\mysurfpoint$ is visible, and $\mykimgsset{\myviewdir}$ are the corresponding viewing directions.
Specifically, ${\myencvec_\myimgk(\mysurfpoint) = \myenc_\myimgk(\mycammat_\myimgk (\myrotmat_\myimgk \mysurfpoint + \mytransvec_\myimgk))}$ using bilinear interpolation.

More formally, the target feature vector for a given 3D surface point $\mysurfpoint$ is computed as
\begin{equation}
    \mytgtvec(\mysurfpoint, \mytgtviewdir) = \myaggrfcn(\mytgtviewdir, \{(\myviewdir_\myimgk, \myencvec_\myimgk(\mysurfpoint))\}_{\myimgk=1}^\mykimgs) \,,
\end{equation}
where $\mykimgs$ is the number of source images that $\mysurfpoint$ is visible in and $\myaggrfcn$ is an aggregation function. The function $\myaggrfcn$ must fulfill a number of criteria; most notably, it should be differentiable and must process any number $K$ of input features, in any order. We explore multiple designs based on differentiable set operators and select one of them based on empirical performance (reported in Section~\ref{sec:experiments}).

A simple choice for $\myaggrfcn$ is a weighted average, where the weights are based on the alignment between the source and target directions:
\begin{equation}
    \myaggrfcn^\text{WA} = \frac{1}{W} \sum_{\myimgk=1}^{\mykimgs} \max(0,\mytgtviewdir^T \myviewdir_\myimgk) \myencvec_\myimgk(\mysurfpoint) \,.
    \label{eqn:aggr_wa}
\end{equation}
Here $W=\sum_{\myimgk=1}^{\mykimgs} \max(0, \mytgtviewdir^T \myviewdir_\myimgk)$ is the sum of all weights.
For a more expressive aggregation function, we can leverage PointNet~\cite{Qi2017PointNet}.
Specifically, we concatenate the source and target directions to the source features, apply an MLP to each feature vector, and aggregate the results:
\begin{equation}
    \myaggrfcn^\text{MLP} = \aggr_{\myimgk=1}^{\mykimgs}  \text{MLP}(\myencvec'_\myimgk) \,.
    \label{eqn:aggr_mlp}
\end{equation}
Here $\myencvec'_\myimgk = [\mytgtviewdir, \myviewdir_\myimgk, \myencvec_\myimgk(\mysurfpoint)]$ is the concatenation of source and target directions with the feature vector, and $\aggr$ is a permutation-invariant operator such as $\mean$ or $\max$.
Instead of an MLP, we can also use a graph attention network (GAT)~\cite{Velivckovic2018GAT} that operates on a fully-connected graph between the source views per 3D point:
\begin{equation}
        \myaggrfcn^\text{GAT} = \aggr_{\myimgk=1}^{\mykimgs}  \text{GAT}\left(\{ \myencvec'_\myimgk \}_{\myimgk=1}^{\mykimgs} \right) \big\vert_\myimgk \,,
    \label{eqn:aggr_gat}
\end{equation}
where $\cdot \vert_\myimgk$ is the readout of the feature vector on node $\myimgk$.

Aggregation functions presented so far compute the target feature $\mytgtvec$ as a set feature.
Another possibility is to read out the target feature vector at the target viewing direction $\mytgtviewdir$.
Specifically, we can create a fully connected graph over source features $\{[\myviewdir_\myimgk, \myencvec_\myimgk]\}_{\myimgk=1}^\mykimgs$ and an initial target feature $[\mytgtviewdir, \mytgtvec']$, where $\mytgtvec'$ is initialized via Equation~\eqref{eqn:aggr_wa}.
Then we can define the readout aggregation function as
\begin{equation}
        \myaggrfcn^\text{GAT-RO} = \text{GAT}\left(\{[\mytgtviewdir, \mytgtvec']\} \cup \{ [\myviewdir_\myimgk, \myencvec_\myimgk(\mysurfpoint)] \}_{\myimgk=1}^{\mykimgs} \right) \big\vert_{0} \,,
    \label{eqn:aggr_gatro}
\end{equation}
where $\cdot \vert_{0}$ denotes the readout of the feature vector associated with the target node.

\boldparagraph{Rendering}
We now describe how the surface points $\mysurfpoint$ are obtained and how the output image $\myout$ in the target view is rendered.
Given a user-specified camera $\mycammat_\mytgtt$ and new camera pose $(\myrotmat_\mytgtt, \mytransvec_\mytgtt)$, we compute a depth map $\mydepthmap \in \myreals^{H \times W}$ from the proxy geometry $\mysurface$.
We then unproject each pixel center of the target view back to 3D based on the depth map $\mydepthmap$, obtaining a surface point for each pixel in $\myout$, $\mysurfpointset$.
Note that $\mydepthmap$ may not have valid depth values for some pixels due to incompleteness of the surface mesh $\mysurface$, or for background regions such as the sky.
We use $\infty$ as the depth value for such pixels.

Given the 3D surface points $\mysurfpointset$, we can compute view-dependent feature vectors $\{\mytgtvec(\mysurfpoint_{h,w})\}_{h,w=1,1}^{H \times W}$ as described above and assemble a feature tensor ${\mytgt = [\mytgtvec_{h,w}]_{h,w=1,1}^{H \times W}}$.
For 3D surface points $\mysurfpoint_{h,w}$ that do not map to any source image, we set $\mytgtvec_{h,w}$ to $\mathbf{0}$.

To synthesize the image $\myout$ from the feature tensor $\mytgt$, we use a convolutional network, denoted by $\myrendernet$: ${\myout = \myrendernet(\mytgt)}$.
The main goal of this network is to regularize the feature map, for example to counteract scale and exposure differences in the source images, and to inpaint missing regions.
For this purpose, we use a sequence of $L$ U-Nets, where each U-Net learns the residual to its input: ${\myrendernet(\mytgt) = \myrendernet^L(\mytgt + \myrendernet^{L-1}(\mytgt + \dots))}$.

\section{Training}
\label{sec:training}

\vspace{-1em}
\boldparagraph{Training a scene-agnostic model}
We train the three networks ($\myencnet$, $\myaggrfcn$, and $\myrendernet$) end-to-end.
Given a set of scenes, we first sample a scene and a source image $\myimg_\myimgn$ that will serve as ground truth.
From the remaining source images of the sampled scene, we sample a subset of $\mymimgs$ source images used for one training pass. %
We then minimize a perceptual loss that is inspired by Chen and Koltun~\cite{Chen2017Photographic}:
\begin{equation}
    \myloss(\myout, \myimg_\myimgn) = ||\myout - \myimg_\myimgn||_1 + \sum_l \lambda_l ||\phi_l(\myout) - \phi_l(\myimg_\myimgn)||_1\,,
\end{equation}
where $\phi_l$ are the outputs of the layers `conv1\_2', `conv2\_2', `conv3\_2', `conv4\_2', and `conv5\_2' of a pretrained VGG-19 network~\cite{Simonyan2015Very}.
We use Adam~\cite{Kingma2014Adam} with a learning rate of $10^{-4}$ and set $\beta_1 = 0.9$, $\beta_2 = 0.9999$, and $\epsilon = 10^{-8}$ to train the network.

\boldparagraph{Network fine-tuning}
The scene-agnostic training procedure described above yields a general network that can be applied to new scenes without retraining or fine-tuning.
However, scenes we apply our method to can be very different from scenes we train on: for example, training the network on Tanks and Temples and applying it on DTU.
We could follow common practice and fine-tune the network parameters $\mathbf{\mynetweights} = [\myencweights, \myaggrweights, \myrenderweights]$ on source images of the target scene, which are provided as input.
Starting from the trained scene-agnostic model, we apply the same training procedure as described above, but only sample training images $\myimg_\myimgn$ from the source images of the target scene.

\boldparagraph{Scene fine-tuning}
An even more powerful form of fine-tuning is to optimize not only the network parameters but also parameters associated with the source images. This enables the optimization to harmonize inconsistencies across images, such as different exposure intervals due to auto-exposure, image-specific motion blur, and other aberrations in the source images.

Recall that so far we have optimized the objective $\min_{\mathbf{\mynetweights}} \myloss(\myout, \myimg_\myimgn)$, where $\mathbf{\mynetweights} = [\myencweights, \myaggrweights, \myrenderweights]$ are the parameters of the encoder, aggregation, and rendering networks.
Note also that the output image $\myout$ produced by the networks is a function of the encoded source images $\{ \myencnet(\myimg_\myimgm; \myencweights) \}_{\myimgm=1}^\mymimgs$.

So far, the image encoder $\myencnet$ took the source image $\myimg_\myimgm$ as input, but the training process only optimized the network parameters $\myencweights$.
The key idea of our more powerful fine-tuning is to also optimize the source images $\{ \myencnet(\myimg_\myimgm; \myencweights) \}_{\myimgm=1}^\mymimgs$ that are used as input. (Importantly, the optimization cannot alter the image $\myimg_\myimgn$ that is used as ground truth in the loss $\myloss(\myout, \myimg_\myimgn)$.)
Specifically, we change the image encoder to $\myencnet(\myimgm; \myencweights, \myimgweights)$, i.e., the input of the network changes from a source image $\myimg_\myimgm$ to the index $\myimgm$, which is used by the network to index into a pool of trainable parameters $\myimgweights$ that are initialized with the actual source images. The source images have become mutable and can be optimized during the training process.
The encoder can also be denoted by $\myencnet(\myimgweights[\myimgm]; \myencweights)$ to establish the connection to the original encoder.

The optimization objective becomes $\min_{\mathbf{\mynetweights}, \myimgweights} \myloss(\myout, \myimg_\myimgn)$.
Aside from the modified objective, the training procedure stays the same.
Note that $\myimgweights$ are initialized with the source images $\{\myimg_\myimgn\}_{\myimgn=1}^\mynimgs$, but the original, unmodified source images $\{\myimg_\myimgn\}_{\myimgn=1}^\mynimgs$ are used throughout the training process in the loss $\myloss(\myout, \myimg_\myimgn)$. Thus the optimization process is forced to produce output $\myout$ that matches the \emph{original} images $\myimg_\myimgn$ and cannot degenerate to a trivial solution such as setting all the source images to a uniform color. The optimization over $\myimgweights$ merely gives the training process the flexibility to modify its perceived input images (e.g., regularizing away inconsistencies) to be able to more closely match the immutable ground-truth targets.

%% file: top_sec_04-eval.tex
\begin{table*}
    \addtolength{\tabcolsep}{-0.2em}
    \centering
    \begin{subtable}[b]{0.3\textwidth}
        \center
        {\scriptsize \input{./tabs/abl_aggr.tex}}
        \caption{3D aggregation function}
        \label{tab:abl_sub_aggr}
    \end{subtable}
    \begin{subtable}[b]{0.3\textwidth}
        \center
        {\scriptsize \input{./tabs/abl_seq.tex}}
        \caption{Number of refinement steps}
        \label{tab:abl_sub_seq}
    \end{subtable}
    \begin{subtable}[b]{0.3\textwidth}
        \center
        {\scriptsize \input{./tabs/abl_ft.tex}}
        \caption{Fine-tuning}
        \label{tab:abl_sub_ft}
    \end{subtable}
    \caption{
        \textbf{Controlled experiments.}
        Mean accuracy over the validation scenes.
        Numbers in bold are within $1\%$ of the best.
    }
    \label{tab:abl}
    \addtolength{\tabcolsep}{0.2em}
\end{table*}

\begin{figure*}
    \center
\begingroup
\setlength{\tabcolsep}{-0.1pt}
\renewcommand{\arraystretch}{1}
\begin{tabular}{c c c}
\raisebox{-0.5\height}{ \includegraphics[width=0.3167\linewidth]{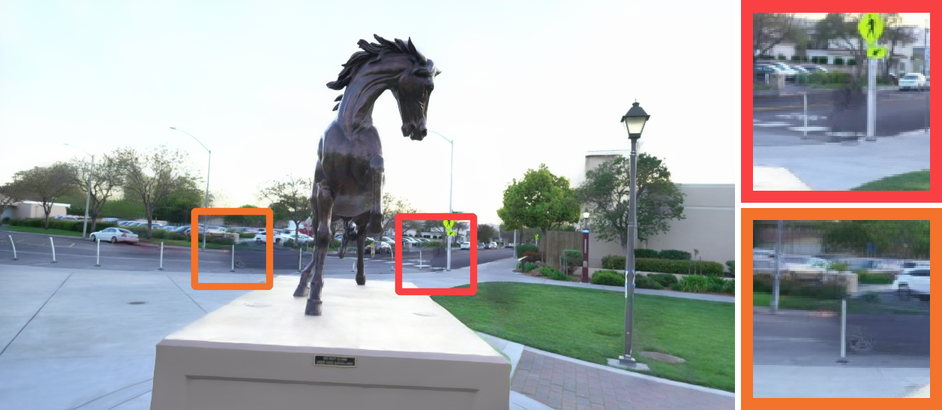} } & \raisebox{-0.5\height}{ \includegraphics[width=0.3167\linewidth]{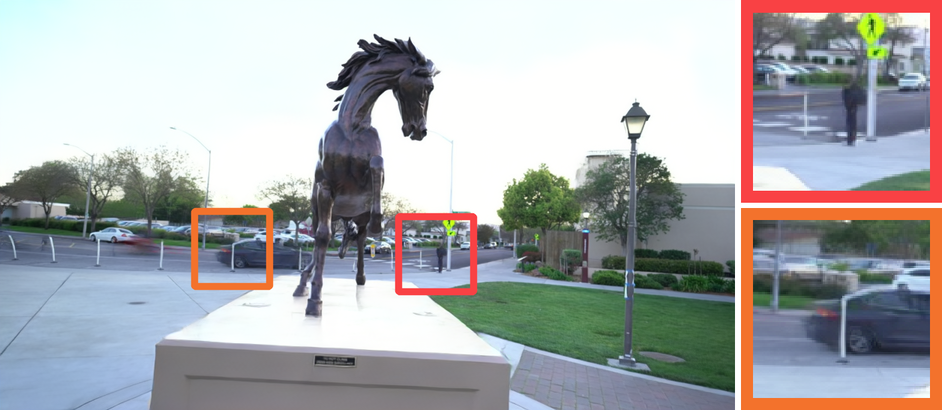} } & \raisebox{-0.5\height}{ \includegraphics[width=0.3167\linewidth]{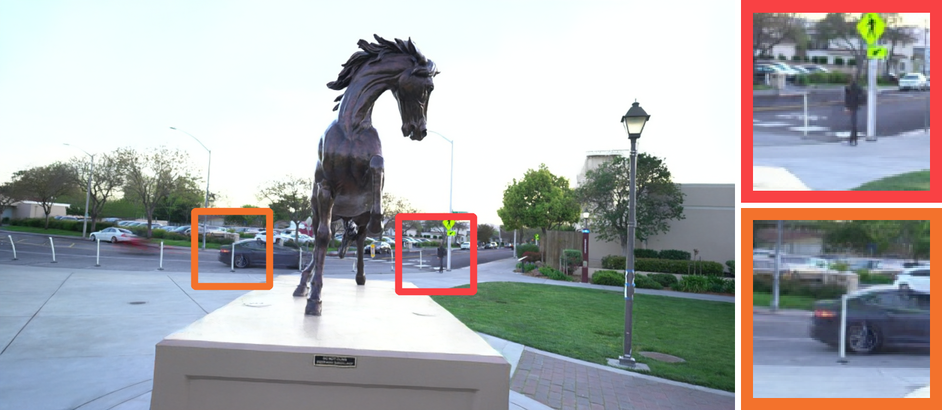} }\vspace{0.25em}\\
General & Network FT & Scene FT
\end{tabular}
\endgroup
    \vspace{-0.5em}
    \caption{
        \textbf{The impact of fine-tuning.} The figure shows a new target view that was not seen by the network during fine-tuning.%
    }
    \label{fig:tat_ft}
    \vspace{-1em}
\end{figure*}

\section{Evaluation}
\label{sec:experiments}

We begin by evaluating our architectural choices in a set of controlled experiments.
We then compare SVS to the state of the art on three challenging datasets: Tanks and Temples~\cite{Knapitsch2017Tanks}, the FVS dataset~\cite{Riegler2020Free}, and DTU~\cite{Aanaes2016DTU}.
We use the same Tanks and Temples scenes for training as Riegler and Koltun~\cite{Riegler2020Free} with the difference that \emph{Ignatius} and \emph{Horse} are withheld for validation, to get a clean split between training, validation, and test scenes.
Thus $15$ of the $21$ Tanks and Temples scenes are used for training, $2$ for validation, and $4$ for evaluation.
We implement the networks in PyTorch~\cite{Paszke2019Pytorch} and train the scene-agnostic model for 600,000 iterations with a batch size of 1, sampling $\mymimgs=3$ source images per iteration.
We use three image fidelity metrics:
LPIPS~\cite{Zhang2018Unreasonable} (reported in percent), which has been shown to correlate well with human perception, alongside SSIM~\cite{Wang2004SSIM} and PSNR, which are metrics that are more attuned to low-level image differences.

\boldparagraph{Architectural choices}
In the first set of controlled experiments, we validate our architectural choices.
As outlined above, we train on $15$ Tanks and Temples scenes and validate on the $2$ withheld scenes. %

First, we compare a set of different 3D aggregation functions.
The results are summarized in Table~\ref{tab:abl_sub_aggr}.
The first row reports the accuracy with the \emph{Weighted Mean} aggregation as described in Equation~\eqref{eqn:aggr_wa}.
The second and third rows report accuracy with the \emph{MLP} aggregation function (see Equation~\eqref{eqn:aggr_mlp}), once with the $\mean$ and once with the $\max$ pooling operator.
Rows four and five report accuracy with the graph attention network aggregation as described in Equation~\eqref{eqn:aggr_gat}, again once with $\mean$ and once with $\max$ pooling of the GAT feature vectors.
The last row reports accuracy with the $\myaggrfcn^\text{GAT-RO}$ aggregation function as defined in Equation~\eqref{eqn:aggr_gatro}.
The results give a slight edge to \emph{MLP Mean} aggregation, in particular for the LPIPS metric, which correlates most reliably with human perception.
We therefore adopt this aggregation function for the other experiments.

In the second experiment, we want to verify that the rendering network benefits from multiple refinement stages. We thus vary the number $L$ of residual U-Net stages in~$\myrendernet$. The results are reported in Table~\ref{tab:abl_sub_seq}.
We observe that there is no significant difference in terms of PSNR and SSIM, but LPIPS decreases with the number of refinement stages.
We thus set $L=9$ for the other experiments.

In the third controlled experiment, we evaluate the impact of scene-specific fine-tuning.
Table~\ref{tab:abl_sub_ft} summarizes the results.
In the first row we show a simple baseline that just averages the RGB values per 3D point and in the second row the network is only trained on the source images of the test scene (not trained on the pre-training scenes).
The third row reports the accuracy of the scene-agnostic network, which is trained on the $15$ training scenes from Tanks and Temples and is not fine-tuned on the validation scenes.
The fourth row reports the accuracy of the same network after fine-tuning the network weights on the source images of the target scene.
(Only the source images are used for fine-tuning. Target views that are used for evaluation are never used during training or fine-tuning.)
The fifth row reports the accuracy of the network after fine-tuning both the network weights and the input images, as described in Section~\ref{sec:training}.
Although none of the fine-tuning methods significantly alters PSNR or SSIM, we can see a clear improvement in LPIPS. We thus use scene fine-tuning for all other experiments.
Figure~\ref{fig:tat_ft} shows the effect of fine-tuning on an example image.

\boldparagraph{Tanks and Temples dataset}
We now compare SVS to the state of the art on four new scenes (not used for training or validation) from the Tanks and Temples dataset~\cite{Knapitsch2017Tanks}, following the protocol of Riegler and Koltun~\cite{Riegler2020Free}.
For each scene, there is a specific set of source images and a disjoint set of target views for evaluation.

\begin{table*}
    \centering
        {\scriptsize \input{./tabs/tat_eval.tex}}
    \vspace{-0.5em}
    \caption{
        \textbf{Accuracy on Tanks and Temples.} Accuracy on the test scenes.
        Numbers in bold are within $1\%$ of the best.
    }
    \label{tab:tat_eval}
\end{table*}

\begin{figure*}
    \center
\begingroup
\setlength{\tabcolsep}{-0.1em}
\renewcommand{\arraystretch}{1}
\begin{tabular}{c c c c c c}
\rotatebox[origin=c]{90}{GT} & \raisebox{-0.5\height}{ \includegraphics[width=0.3167\linewidth]{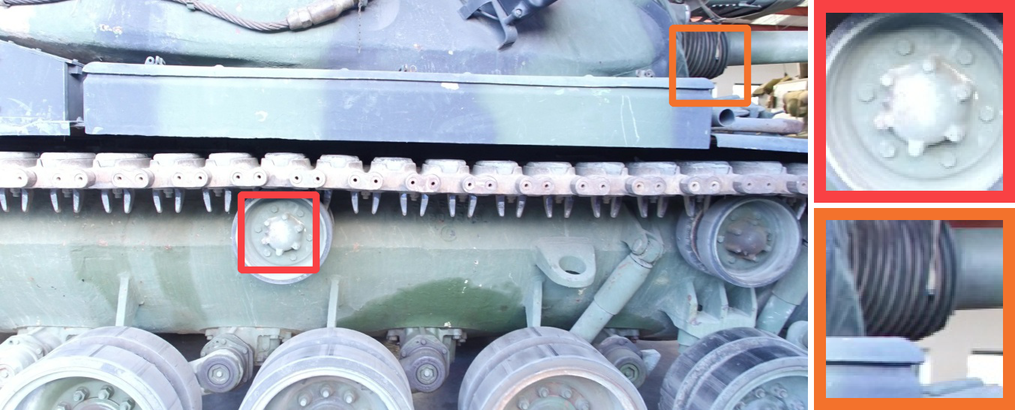} } & \raisebox{-0.5\height}{ \includegraphics[width=0.3167\linewidth]{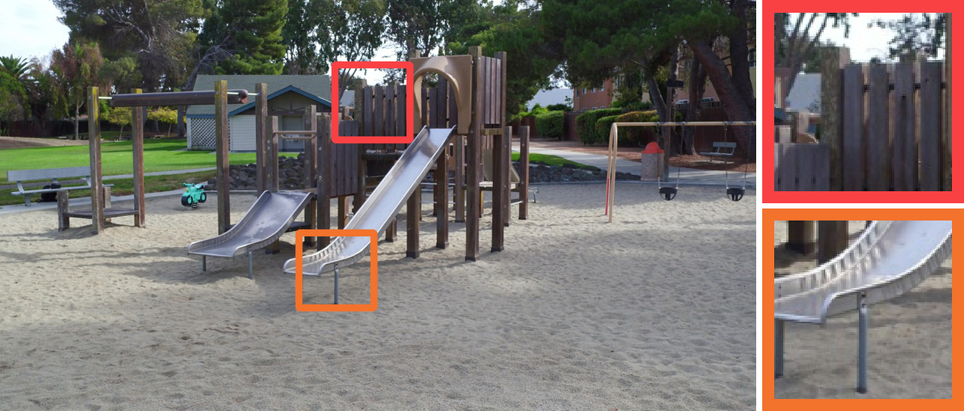} } & \raisebox{-0.5\height}{ \includegraphics[width=0.3167\linewidth]{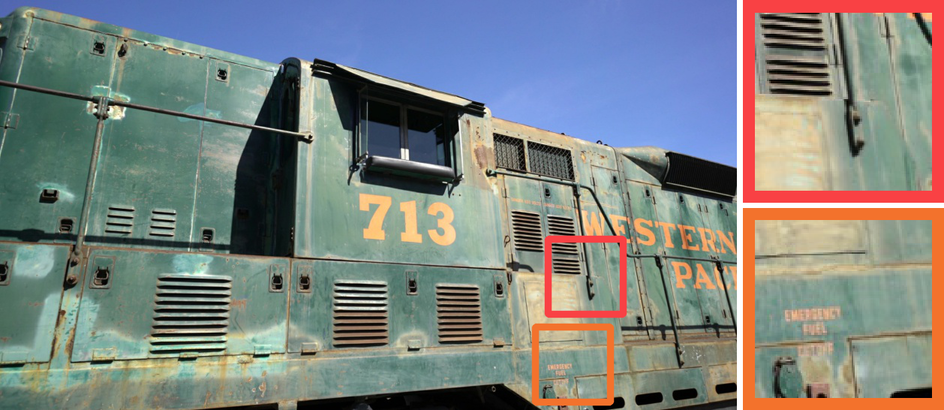} }\vspace{0.25em}\\
\rotatebox[origin=c]{90}{Ours} & \raisebox{-0.5\height}{ \includegraphics[width=0.3167\linewidth]{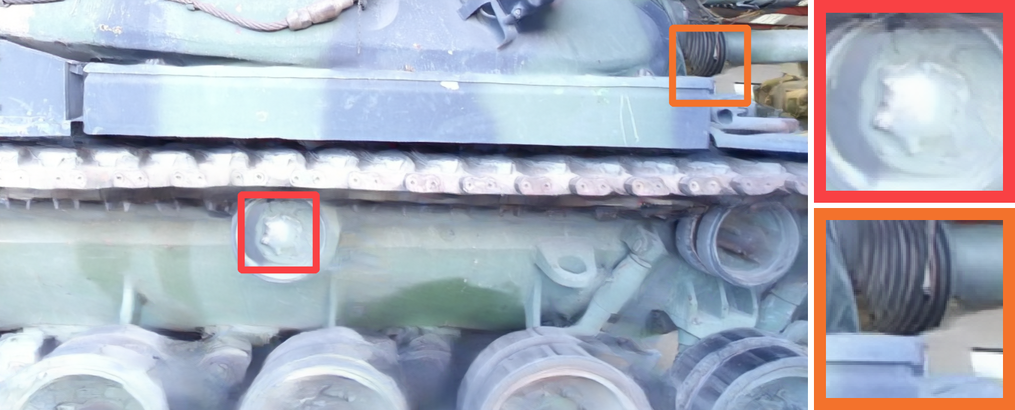} } & \raisebox{-0.5\height}{ \includegraphics[width=0.3167\linewidth]{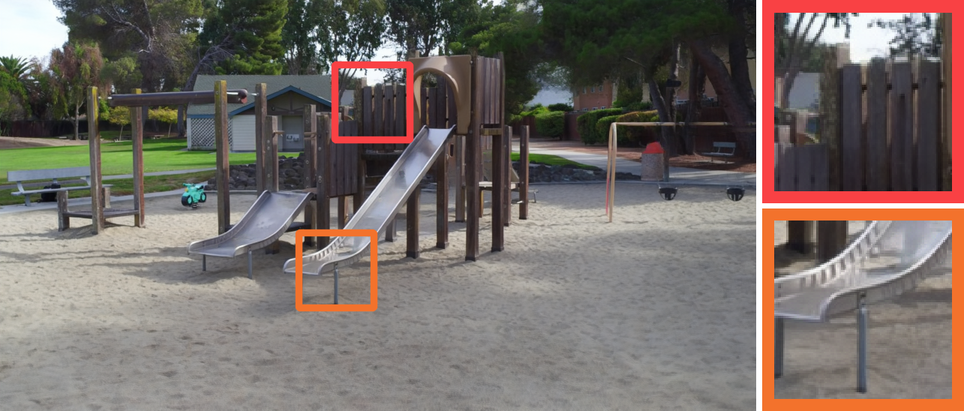} } & \raisebox{-0.5\height}{ \includegraphics[width=0.3167\linewidth]{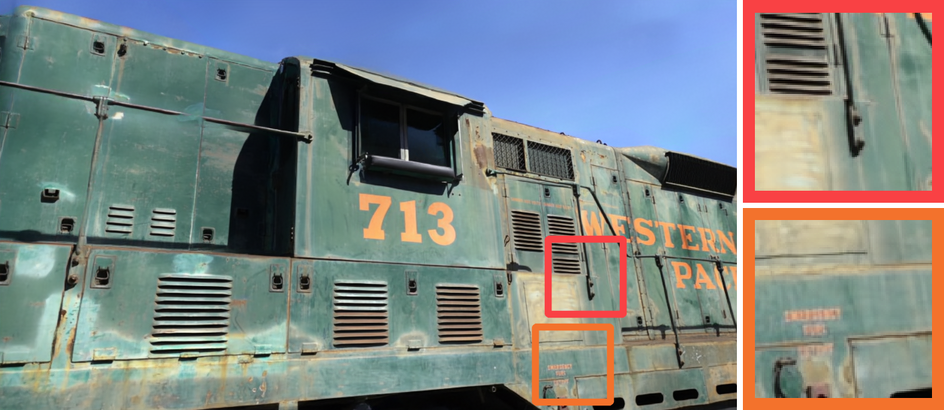} }\vspace{0.25em}\\
\rotatebox[origin=c]{90}{FVS \cite{Riegler2020Free}} & \raisebox{-0.5\height}{ \includegraphics[width=0.3167\linewidth]{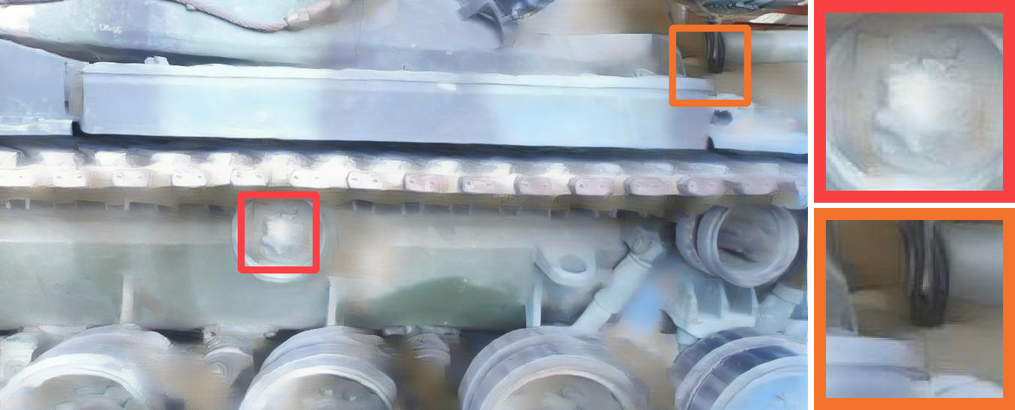} } & \raisebox{-0.5\height}{ \includegraphics[width=0.3167\linewidth]{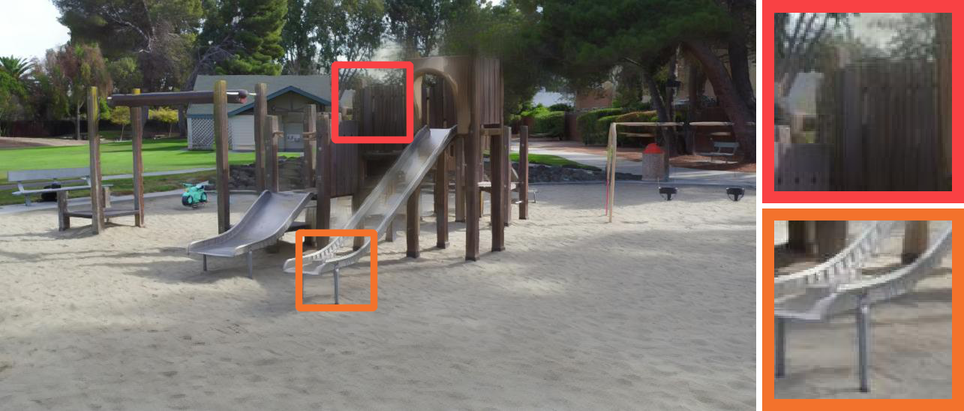} } & \raisebox{-0.5\height}{ \includegraphics[width=0.3167\linewidth]{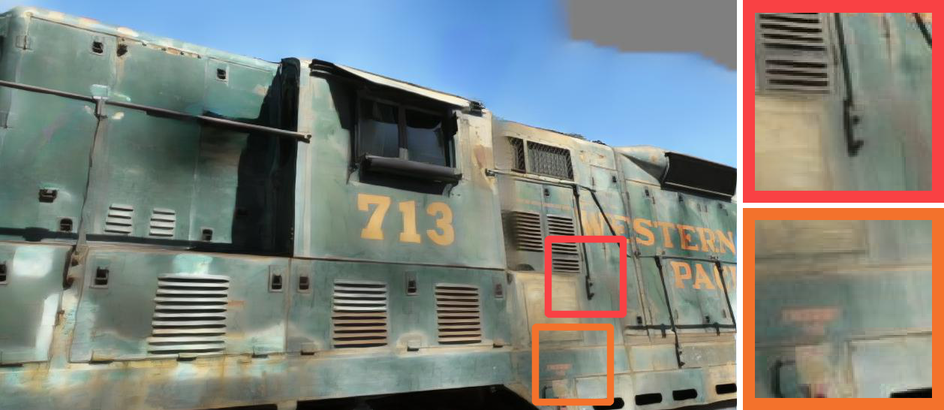} }\vspace{0.25em}\\
\rotatebox[origin=c]{90}{NeRF++ \cite{Zhang2020Nerfpp}} & \raisebox{-0.5\height}{ \includegraphics[width=0.3167\linewidth]{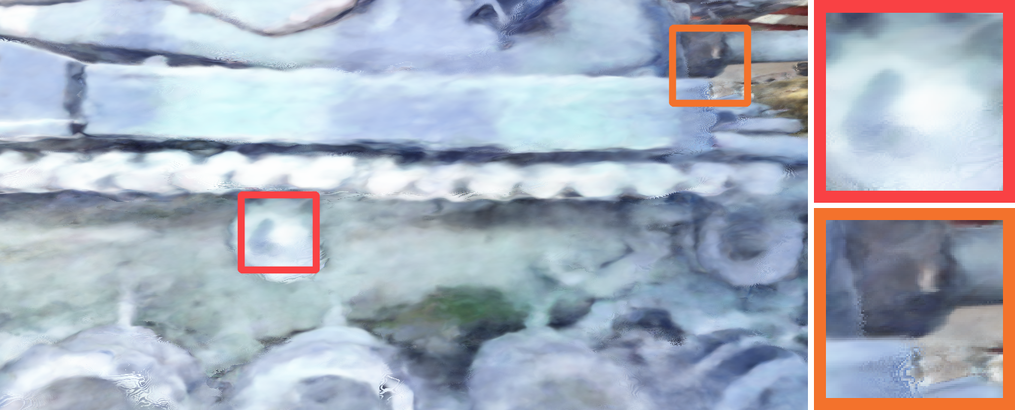} } & \raisebox{-0.5\height}{ \includegraphics[width=0.3167\linewidth]{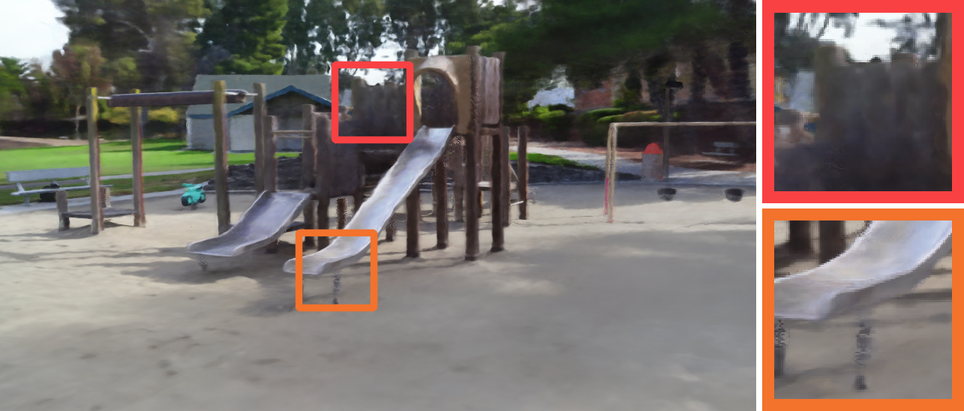} } & \raisebox{-0.5\height}{ \includegraphics[width=0.3167\linewidth]{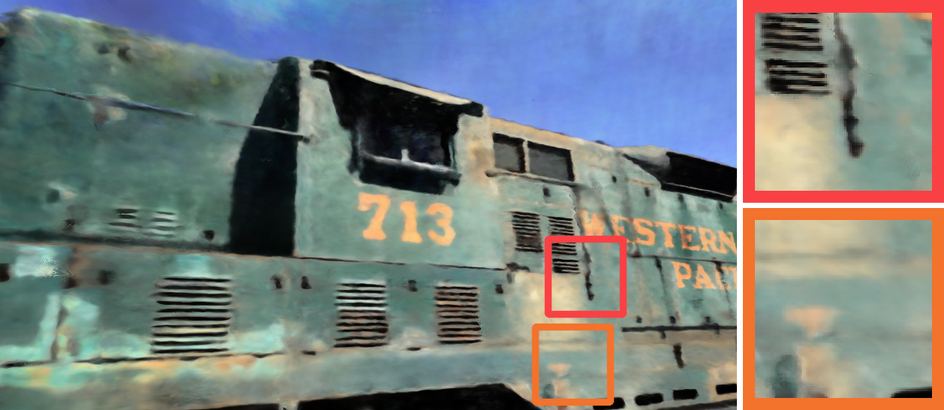} }\vspace{0.25em}\\
\rotatebox[origin=c]{90}{NPBG \cite{Aliev2020Neural}} & \raisebox{-0.5\height}{ \includegraphics[width=0.3167\linewidth]{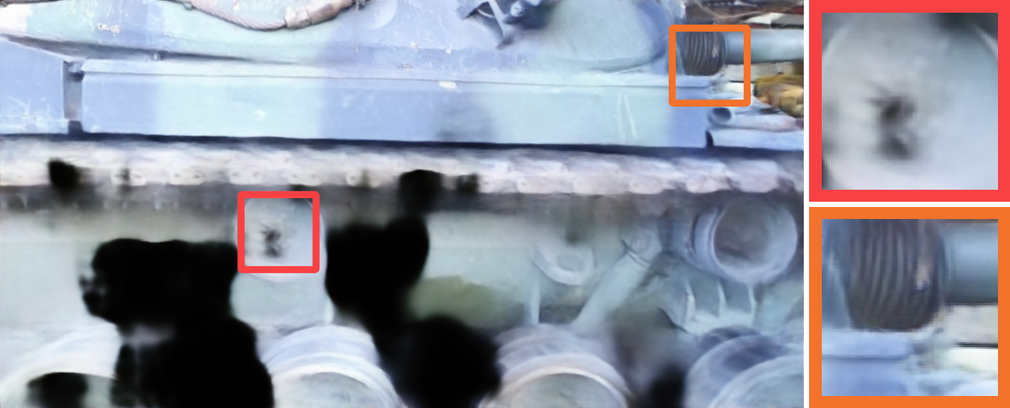} } & \raisebox{-0.5\height}{ \includegraphics[width=0.3167\linewidth]{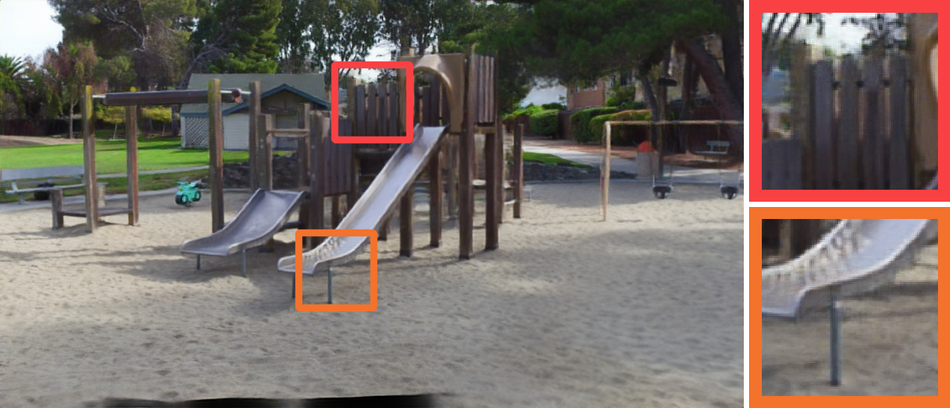} } & \raisebox{-0.5\height}{ \includegraphics[width=0.3167\linewidth]{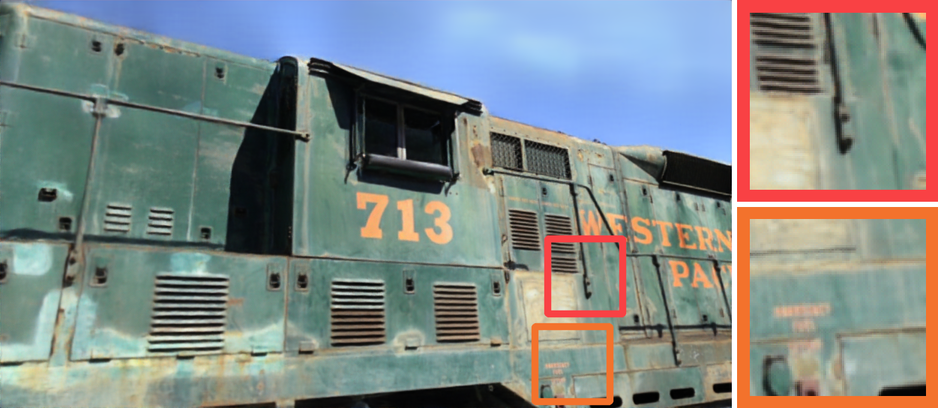} }\vspace{0.25em}\\
 & M60 & Playground & Train\\
\end{tabular}
\endgroup
    \vspace{-1em}
    \caption{
        \textbf{Qualitative results on Tanks and Temples.}
        Comparison of SVS to the best-performing prior methods.
    }
    \label{fig:tat_eval}
    \vspace{-1em}
\end{figure*}

\begin{table*}
    \centering
    \resizebox{1\linewidth}{!}{
    {\scriptsize \input{./tabs/fvs.tex} }
    }
    \vspace{-0.5em}
    \caption{
        \textbf{Accuracy on the FVS dataset.}
        Numbers in bold are within $1\%$ of the best.
    }
    \label{tab:fvs}
\end{table*}

\begin{table*}
    \centering
    {\scriptsize \input{./tabs/dtu_eval.tex} }
    \vspace{-0.5em}
    \caption{
        \textbf{Accuracy on DTU.}
        Numbers in bold are within $1\%$ of the best.
        In each column, numbers on the left are for view interpolation, right for extrapolation.
    }
    \label{tab:dtu}
    \vspace{-1em}
\end{table*}

We compare to a variety of recent methods that represent different approaches to view synthesis and have been applied in comparable settings in the past.
For Local Light Field Fusion~(\emph{LLFF})~\cite{Mildenhall2019Local} we used the publicly available code.
Since no training code is available, we use the provided pretrained network weights.
For Extreme View Synthesis~(\emph{EVS})~\cite{Choi2019Extreme} we also use the publicly available code and the provided network weights.
Neural Point Based Graphics~(\emph{NPBG})~\cite{Aliev2020Neural} is fitted per scene using the published code and pretrained rendering network weights.
For Neural Radiance Fields~(\emph{NeRF})~\cite{Mildenhall2020Nerf} and \emph{NeRF++}~\cite{Zhang2020Nerfpp} we manually define the bounding volume around the main object in each scene.
These approaches are trained per scene.
For Free View Synthesis~(\emph{FVS})~\cite{Riegler2020Free} we use the publicly available code and the published network weights, which had been trained on the union of our training and validation scenes.

The results are summarized in Table~\ref{tab:tat_eval}.
As observed in prior work~\cite{Riegler2020Free}, \emph{LLFF} and \emph{EVS} struggle in this challenging view synthesis setting.
We also see that \emph{NeRF++} improves over \emph{NeRF}, but neither attain the accuracy of the best-performing methods.
SVS without any scene-specific fine-tuning (\emph{Ours w/o FT}) already outperforms all prior work for most scenes, especially with respect to LPIPS. Our full method (\emph{Ours})
achieves the best results on all scenes.

Figure~\ref{fig:tat_eval} shows images synthesized by the best-performing methods on a number of scenes.
\emph{FVS} sometimes fails to utilize all the relevant images, which leads to missing regions.
\emph{NeRF++} suffers from blurring and patterning in the output, although it sometimes reconstructs details that are missing in our geometric scaffold.
While the results of \emph{NPBG} can be very good, it sometimes introduces noticeable artifacts in parts of the scene.
Images synthesized by SVS are overall sharper, more complete, more accurate, and more temporally stable than the prior work. Please see the supplementary video for sequences.

\boldparagraph{Free View Synthesis dataset}
Next, we compare SVS with prior work on the FVS dataset~\cite{Riegler2020Free}.
This dataset contains $6$ scenes, each of which was recorded at least twice.
The first recording provides the source images and the other recordings serve as ground truth for novel target views.
Quantitative results are summarized in Table~\ref{tab:fvs} and qualitative results are provided in the supplement.
Due to space constraints, we omit PSNR values here.
SVS improves over prior work on all scenes, according to all metrics. Note that SVS reduces the LPIPS relative to the best prior method by at least $5$ absolute percentage points in every scene.

\boldparagraph{DTU}
Lastly, we compare SVS to prior approaches on the DTU dataset~\cite{Aanaes2016DTU}.
DTU scenes are captured with a regular camera layout, where $49$ images are taken from an octant of a sphere.
We follow the protocol of Riegler and Koltun~\cite{Riegler2020Free}, use the same scenes, and use the $6$ central cameras to evaluate view interpolation and the $4$ corner cameras to evaluate view extrapolation.

Quantitative results are summarized in Table~\ref{tab:dtu} and qualitative results are provided in the supplement.
\emph{LLFF} and \emph{EVS} achieve reasonable results on this dataset, indicating that this setup
conforms much better to their modeling assumptions.
\emph{NPBG} struggles on this dataset, possibly due to the small number of images per scene (i.e., 39). %
\emph{NeRF} excels on this dataset;
we manually specified a tight bounding box around the object to maximize the accuracy of \emph{NeRF}.
The results of \emph{FVS} are on par with \emph{NeRF} with respect to SSIM and LPIPS.
For our method, the scene-agnostic model, which was trained on Tanks and Temples and has never seen DTU-like scenes, is already surprisingly competitive, and the full SVS method sets the new state of the art for novel view synthesis on this dataset with respect to LPIPS, attaining an average LPIPS error of 4.5\% in extrapolation mode and 1.6\% for view interpolation.

%% file: tabs/abl_aggr.tex
\begin{tabular}{rrrr}
\toprule
 & $\uparrow$PSNR & $\uparrow$SSIM & $\downarrow$LPIPS$\%$ \\
\midrule
\multicolumn{1}{l}{{\color[rgb]{0,0,0} Weighted Mean}} & {\bf 21.42} & {\bf 0.870} & 12.84 \\
\multicolumn{1}{l}{{\color[rgb]{0,0,0} MLP Mean}} & {\bf 21.25} & {\bf 0.869} & {\bf 12.51} \\
\multicolumn{1}{l}{{\color[rgb]{0,0,0} MLP Max}} & 20.95 & {\bf 0.863} & 12.65 \\
\multicolumn{1}{l}{{\color[rgb]{0,0,0} GAT Mean}} & 21.01 & {\bf 0.864} & 12.84 \\
\multicolumn{1}{l}{{\color[rgb]{0,0,0} GAT Max}} & 21.05 & {\bf 0.864} & 13.09 \\
\multicolumn{1}{l}{{\color[rgb]{0,0,0} GAT Readout}} & 20.88 & {\bf 0.862} & 12.81 \\
\bottomrule
\end{tabular}

%% file: tabs/abl_seq.tex
\begin{tabular}{rrrr}
\toprule
 & $\uparrow$PSNR & $\uparrow$SSIM & $\downarrow$LPIPS$\%$ \\
\midrule
\multicolumn{1}{l}{{\color[rgb]{0,0,0} 1}} & 21.20 & {\bf 0.868} & 12.62 \\
\multicolumn{1}{l}{{\color[rgb]{0,0,0} 3}} & 21.25 & {\bf 0.869} & 12.51 \\
\multicolumn{1}{l}{{\color[rgb]{0,0,0} 5}} & 21.30 & {\bf 0.870} & 12.46 \\
\multicolumn{1}{l}{{\color[rgb]{0,0,0} 7}} & {\bf 21.55} & {\bf 0.872} & 12.41 \\
\multicolumn{1}{l}{{\color[rgb]{0,0,0} 9}} & {\bf 21.39} & {\bf 0.871} & {\bf 12.27} \\
\bottomrule
\end{tabular}

%% file: tabs/abl_ft.tex
\begin{tabular}{rrrr}
\toprule
 & $\uparrow$PSNR & $\uparrow$SSIM & $\downarrow$LPIPS$\%$ \\
\midrule
\multicolumn{1}{l}{RGB Averaging} & 21.15 & 0.844 & 22.84 \\
\multicolumn{1}{l}{Network FT w/o PT} & 21.13 & 0.865 & 15.05 \\
\multicolumn{1}{l}{{\color[rgb]{0,0,0} General}} & 21.59 & {\bf 0.872} & 12.19 \\
\multicolumn{1}{l}{{\color[rgb]{0,0,0} Network FT}} & {\bf 22.16} & {\bf 0.874} & 11.26 \\
\multicolumn{1}{l}{{\color[rgb]{0,0,0} Scene FT}} & {\bf 22.02} & {\bf 0.873} & {\bf 9.99} \\
\bottomrule
\end{tabular}

%% file: tabs/tat_eval.tex
\begin{tabular}{rrrrrrrrrrrrr}
\toprule
\multicolumn{1}{c}{} & \multicolumn{3}{c}{Truck} & \multicolumn{3}{c}{M60} & \multicolumn{3}{c}{Playground} & \multicolumn{3}{c}{Train} \\
\cmidrule(lr){2-4}
\cmidrule(lr){5-7}
\cmidrule(lr){8-10}
\cmidrule(lr){11-13}
 & $\uparrow$PSNR & $\uparrow$SSIM & $\downarrow$LPIPS$\%$ & $\uparrow$PSNR & $\uparrow$SSIM & $\downarrow$LPIPS$\%$ & $\uparrow$PSNR & $\uparrow$SSIM & $\downarrow$LPIPS$\%$ & $\uparrow$PSNR & $\uparrow$SSIM & $\downarrow$LPIPS$\%$ \\
\midrule
\multicolumn{1}{l}{{\color[rgb]{0,0,0} LLFF \cite{Mildenhall2019Local}}} & 10.78 & 0.454 & 60.62 & 8.98 & 0.431 & 71.76 & 14.40 & 0.578 & 53.93 & 9.15 & 0.384 & 67.40 \\
\multicolumn{1}{l}{{\color[rgb]{0,0,0} EVS \cite{Choi2019Extreme}}} & 14.22 & 0.527 & 43.52 & 7.41 & 0.354 & 75.71 & 14.72 & 0.568 & 46.85 & 10.54 & 0.378 & 67.62 \\
\multicolumn{1}{l}{{\color[rgb]{0,0,0} NPBG \cite{Aliev2020Neural}}} & 21.88 & 0.877 & 15.04 & 12.35 & 0.716 & 35.57 & 23.03 & {\bf 0.876} & 16.65 & 18.08 & 0.801 & 25.48 \\
\multicolumn{1}{l}{{\color[rgb]{0,0,0} NeRF \cite{Mildenhall2020Nerf}}} & 20.85 & 0.738 & 50.74 & 16.86 & 0.701 & 60.89 & 21.55 & 0.759 & 52.19 & 16.64 & 0.627 & 64.64 \\
\multicolumn{1}{l}{{\color[rgb]{0,0,0} NeRF++ \cite{Zhang2020Nerfpp}}} & 22.77 & 0.814 & 30.04 & 18.49 & 0.747 & 43.06 & 22.93 & 0.806 & 38.70 & 17.77 & 0.681 & 47.75 \\
\multicolumn{1}{l}{{\color[rgb]{0,0,0} FVS \cite{Riegler2020Free}}} & 22.93 & 0.873 & 13.06 & 16.83 & 0.783 & 30.70 & 22.28 & 0.846 & 19.47 & 18.09 & 0.773 & 24.74 \\
\midrule
\multicolumn{1}{l}{{\color[rgb]{0,0,0} Ours w/o FT}} & 23.09 & {\bf 0.893} & 12.41 & 19.41 & {\bf 0.827} & 23.70 & {\bf 23.61} & {\bf 0.876} & 17.38 & 18.42 & 0.809 & 19.42 \\
\multicolumn{1}{l}{{\color[rgb]{0,0,0} Ours}} & {\bf 23.86} & {\bf 0.895} & {\bf 9.34} & {\bf 19.97} & {\bf 0.833} & {\bf 20.45} & {\bf 23.72} & {\bf 0.884} & {\bf 14.22} & {\bf 18.69} & {\bf 0.820} & {\bf 15.73} \\
\bottomrule
\end{tabular}

%% file: tabs/fvs.tex
\begin{tabular}{rrrrrrrrrrrrr}
\toprule
\multicolumn{1}{c}{} & \multicolumn{2}{c}{Bike} & \multicolumn{2}{c}{Flowers} & \multicolumn{2}{c}{Pirate} & \multicolumn{2}{c}{Digger} & \multicolumn{2}{c}{Sandbox} & \multicolumn{2}{c}{Soccertable} \\
\cmidrule(lr){2-3}
\cmidrule(lr){4-5}
\cmidrule(lr){6-7}
\cmidrule(lr){8-9}
\cmidrule(lr){10-11}
\cmidrule(lr){12-13}
 & $\uparrow$SSIM & $\downarrow$LPIPS$\%$ & $\uparrow$SSIM & $\downarrow$LPIPS$\%$ & $\uparrow$SSIM & $\downarrow$LPIPS$\%$ & $\uparrow$SSIM & $\downarrow$LPIPS$\%$ & $\uparrow$SSIM & $\downarrow$LPIPS$\%$ & $\uparrow$SSIM & $\downarrow$LPIPS$\%$ \\
\midrule
\multicolumn{1}{l}{{\color[rgb]{0,0,0} NPBG \cite{Aliev2020Neural}}} & 0.616 & 31.08 & 0.553 & 48.47 & 0.592 & 45.71 & 0.686 & 29.08 & 0.650 & 35.91 & 0.723 & 29.97 \\
\multicolumn{1}{l}{{\color[rgb]{0,0,0} NeRF++ \cite{Zhang2020Nerfpp}}} & 0.715 & 27.01 & 0.816 & 30.30 & 0.712 & 41.56 & 0.657 & 34.69 & 0.842 & 23.08 & 0.889 & 20.61 \\
\multicolumn{1}{l}{{\color[rgb]{0,0,0} FVS \cite{Riegler2020Free}}} & 0.592 & 27.83 & 0.778 & 26.07 & 0.685 & 35.89 & 0.668 & 23.27 & 0.770 & 30.20 & 0.819 & 19.41 \\
\midrule
\multicolumn{1}{l}{{\color[rgb]{0,0,0} Ours w/o FT}} & 0.745 & 21.18 & {\bf 0.848} & 21.41 & 0.752 & {\bf 29.21} & 0.782 & 18.00 & 0.850 & 21.48 & 0.895 & 14.79 \\
\multicolumn{1}{l}{{\color[rgb]{0,0,0} Ours}} & {\bf 0.757} & {\bf 20.84} & {\bf 0.845} & {\bf 20.82} & {\bf 0.760} & 30.83 & {\bf 0.791} & {\bf 16.12} & {\bf 0.862} & {\bf 20.00} & {\bf 0.912} & {\bf 13.07} \\
\bottomrule
\end{tabular}

%% file: tabs/dtu_eval.tex
\begin{tabular}{rrrrrrrrrr}
\toprule
\multicolumn{1}{c}{} & \multicolumn{3}{c}{65} & \multicolumn{3}{c}{106} & \multicolumn{3}{c}{118} \\
\cmidrule(lr){2-4}
\cmidrule(lr){5-7}
\cmidrule(lr){8-10}
 & \multicolumn{1}{c}{$\uparrow$PSNR} & \multicolumn{1}{c}{$\uparrow$SSIM} & \multicolumn{1}{c}{$\downarrow$LPIPS$\%$} & \multicolumn{1}{c}{$\uparrow$PSNR} & \multicolumn{1}{c}{$\uparrow$SSIM} & \multicolumn{1}{c}{$\downarrow$LPIPS$\%$} & \multicolumn{1}{c}{$\uparrow$PSNR} & \multicolumn{1}{c}{$\uparrow$SSIM} & \multicolumn{1}{c}{$\downarrow$LPIPS$\%$} \\
\midrule
\multicolumn{1}{l}{LLFF \cite{Mildenhall2019Local}} & 22.48/22.07 & 0.935/0.921 & 9.38/12.71 & 24.10/24.63 & 0.900/0.886 & 13.26/13.57 & 28.99/27.42 & 0.928/0.922 & 9.69/10.99 \\
\multicolumn{1}{l}{EVS \cite{Choi2019Extreme}} & 23.26/14.43 & 0.942/0.848 & 7.94/22.11 & 20.21/11.15 & 0.902/0.743 & 14.91/29.57 & 23.35/12.06 & 0.928/0.793 & 10.84/25.01 \\
\multicolumn{1}{l}{NPBG \cite{Aliev2020Neural}} & 16.74/15.44 & 0.889/0.873 & 14.30/19.45 & 19.62/20.26 & 0.847/0.842 & 18.90/21.13 & 23.81/24.14 & 0.867/0.879 & 15.22/16.88 \\
\multicolumn{1}{l}{NeRF \cite{Mildenhall2020Nerf}} & {\bf 32.00}/{\bf 28.12} & {\bf 0.984}/{\bf 0.963} & 3.04/8.54 & {\bf 34.45}/{\bf 30.66} & {\bf 0.975}/{\bf 0.957} & 7.02/10.14 & {\bf 37.36}/{\bf 31.66} & {\bf 0.985}/{\bf 0.967} & 4.18/6.92 \\
\multicolumn{1}{l}{FVS \cite{Riegler2020Free}} & 30.44/25.32 & {\bf 0.984}/{\bf 0.961} & 2.56/7.17 & 32.96/27.56 & {\bf 0.979}/0.950 & 2.96/6.57 & 35.64/29.54 & {\bf 0.985}/{\bf 0.963} & 1.95/6.31 \\
\midrule
\multicolumn{1}{l}{Ours w/o FT} & 30.08/23.98 & {\bf 0.983}/{\bf 0.960} & 2.36/7.16 & 32.06/29.01 & {\bf 0.978}/{\bf 0.959} & 3.54/5.36 & 35.65/30.42 & {\bf 0.986}/{\bf 0.966} & 2.15/5.15 \\
\multicolumn{1}{l}{Ours} & {\bf 32.13}/26.82 & {\bf 0.986}/{\bf 0.964} & {\bf 1.70}/{\bf 5.61} & {\bf 34.30}/{\bf 30.64} & {\bf 0.983}/{\bf 0.965} & {\bf 1.93}/{\bf 3.69} & {\bf 37.27}/{\bf 31.44} & {\bf 0.988}/{\bf 0.967} & {\bf 1.30}/{\bf 4.26} \\
\bottomrule
\end{tabular}

%% file: top_sec_05-conclusion.tex
\section{Discussion}
We presented a view synthesis method that is based on differentiable on-surface feature processing.
The method aggregates deep features from source images adaptively on a geometric scaffold of the scene using a differentiable set network. The pipeline is trained end-to-end and learns to aggregate features from all images, obviating the need for heuristic selection of ``relevant'' source images.
Our method sets a new state of the art for photorealistic view synthesis on large-scale real-world scenes.

There are a number of exciting avenues for future work.
First, we look forward to continued progress in 3D reconstruction~\cite{Knapitsch2017Tanks}, which can further advance the fidelity of the images synthesized by the presented approach.
Second, it would be interesting to extend the approach to image sets with strong appearance variation, perhaps enabling relighting of the scenes at test time~\cite{Li2020Crowdsampling, Martin2020NeRFW}.
Lastly, the presented approach, like most recent view synthesis work, only handles static scenes. This enables the user to look at these environments but not engage and interact with them. An exciting challenge for the field is to enable interactive manipulation of such scenes while maintaining photorealism.

%% file: top_sec_06-appendix.tex
\mysuppsection{Algorithmic Overview and Implementation}
Our Stable View Synthesis method has two major stages.
In a first stage we set up the scene from a set of input images as outlined in Algorithm~\ref{alg:setup}.
This includes erecting the geometric scaffold and encoding all source images.
In the second stage, we actually synthesize new images from novel target views.
Given the preprocessed scene and a user-specified camera and viewpoint, we synthesize the new image as outlined in Algorithm~\ref{alg:svs}.

Note that the steps in Algorithm~\ref{alg:svs} are easily parallelizable.
For each pixel in the target view we backproject a single 3D point, which can be efficiently implemented as a single matrix-matrix multiplication to unproject all pixels.
For the aggregation in 3D we first have to project each 3D point into each source image.
If it projects to a valid image location, we bilinearly interpolate the feature vector.
These operations are trivially parallelizable over the 3D points.
The aggregation function itself can then be efficiently implemented with optimized gather operations as implemented in PyTorch Geometric~\cite{Fey2019PyTorchGeometric}.
The concatenation of 3D feature vectors and synthesizing an output image are implemented with standard functions and building blocks of the deep learning framework~\cite{Paszke2019Pytorch}.

\mysuppsection{Evaluation Details}
We train the scene-agnostic model with 600,000 iterations.
The scene-specific model is trained with $256 \cdot \mynimgs$ iterations starting from the scene-agnostic model, where $\mynimgs$ is the number of source images for the given scene.

The scene-agnostic model is trained on quarter-resolution images from the Tanks and Temples dataset~\cite{Knapitsch2017Tanks}.
To be comparable to prior work, we evaluate our method on half-resolution images.
For Tanks and Temples~\cite{Knapitsch2017Tanks} the output images are $980 \times 546$ pixels (slight variations for different scenes), for the FVS dataset~\cite{Riegler2020Free} the images are $990 \times 543$ pixels (slight variations for different scenes), and on DTU the images are $800 \times 600$ pixels.

\algrenewcommand{\algorithmiccomment}[1]{$\vartriangleright$\emph{ #1}}
\begin{algorithm}[H]
    \begin{algorithmic}[1]
        \Statex \Comment{Erect geometric scaffold}
        \State $\{\mycammat_\myimgn, \myrotmat_\myimgn, \mytransvec_\myimgn\}_{\myimgn=1}^\mynimgs = \text{structure-from-motion}(\myimgset)$
        \State $\mysurface = \text{surface-reconstruction}(\{\myimg_\myimgn, \mycammat_\myimgn, \myrotmat_\myimgn, \mytransvec_\myimgn\}_{\myimgn=1}^\mynimgs)$

        \Statex
        \Statex \Comment{Encode source}
        \ForAll{$\myimg_\myimgn$ in $\myimgset$}
            \State $\myenc_\myimgn = \myencnet(\myimg_\myimgn)$
        \EndFor
    \end{algorithmic}
    \caption{
        \textbf{Scene Setup.}
        Input is a set of source images $\myimgset$ and the outputs are source camera parameters and viewpoints $\{\mycammat_\myimgn, \myrotmat_\myimgn, \mytransvec_\myimgn\}_{\myimgn=1}^\mynimgs$, the geometric scaffold $\mysurface$, and the encoded source images $\{\myenc_\myimgn\}_{\myimgn=1}^\mynimgs$.
    }
    \label{alg:setup}
\end{algorithm}

\begin{algorithm}[H]
    \begin{algorithmic}[1]
        \Statex \Comment{Get surface points}
        \State $D = \text{render}(\mysurface, \mycammat_\mytgtt, \myrotmat_\mytgtt, \mytransvec_\mytgtt)$
        \State $\mysurfpoints = \text{unproject}(D, \mycammat_\mytgtt, \myrotmat_\mytgtt, \mytransvec_\mytgtt)$

        \Statex
        \Statex \Comment{Aggregate feature vectors per 3D point}
        \ForAll{$\mysurfpoint_{h,w}$ in $\mysurfpoints$}
        \State $\mytgtvec((\mysurfpoint, \mytgtviewdir)_{h,w}) = \myaggrfcn(\mytgtviewdir_{h,w}, \{(\myviewdir_\myimgk, \myencvec_\myimgk(\mysurfpoint_{h,w}))\}_{\myimgk=1}^\mykimgs)$
        \EndFor

        \Statex
        \Statex \Comment{Render Image}
        \State $\mytgt = [\mytgtvec((\mysurfpoint, \mytgtviewdir)_{h,w})]_{h,w=1,1}^{H \times W}$
        \State $O = \myrendernet(\mytgt)$
    \end{algorithmic}
    \caption{
        \textbf{Stable View Synthesis.}
        Input is the preprocessed scene $(\{\mycammat_\myimgn, \myrotmat_\myimgn, \mytransvec_\myimgn\}_{\myimgn=1}^\mynimgs, \mysurface, \{\myenc_\myimgn\}_{\myimgn=1}^\mynimgs)$ and a target view defined by camera matrix $\mycammat_\mytgtt$ and pose $\myrotmat_\mytgtt, \mytransvec_\mytgtt$, output is an image $\myout$ of the scene in the target view.
    }
    \label{alg:svs}
\end{algorithm}

\mysuppsection{Additional Results}

\begin{figure*}
    \center
    \begingroup
\setlength{\tabcolsep}{-0.1em}
\renewcommand{\arraystretch}{1}
\begin{tabular}{c c c c c c}
\rotatebox[origin=c]{90}{GT} & \raisebox{-0.5\height}{ \includegraphics[width=0.3167\linewidth]{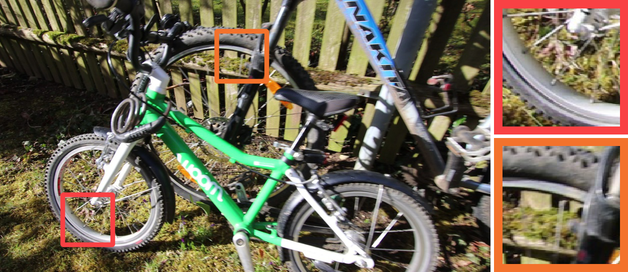} } & \raisebox{-0.5\height}{ \includegraphics[width=0.3167\linewidth]{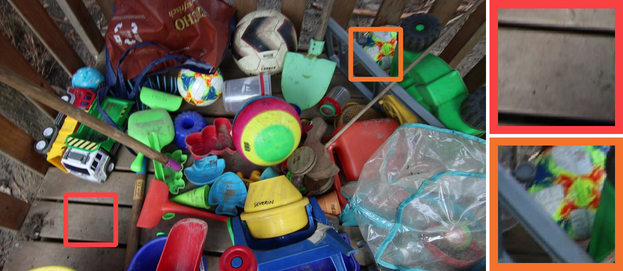} } & \raisebox{-0.5\height}{ \includegraphics[width=0.3167\linewidth]{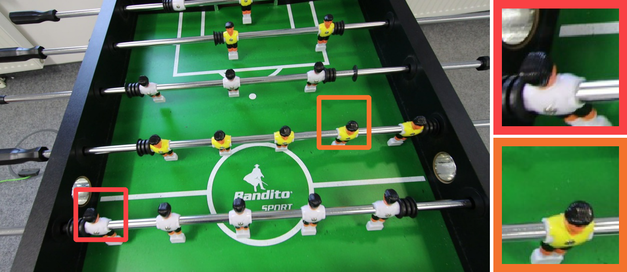} }\vspace{0.25em}\\
\rotatebox[origin=c]{90}{Ours} & \raisebox{-0.5\height}{ \includegraphics[width=0.3167\linewidth]{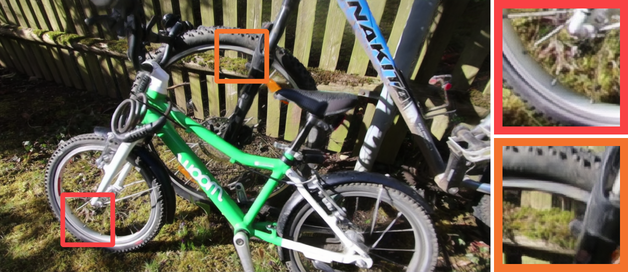} } & \raisebox{-0.5\height}{ \includegraphics[width=0.3167\linewidth]{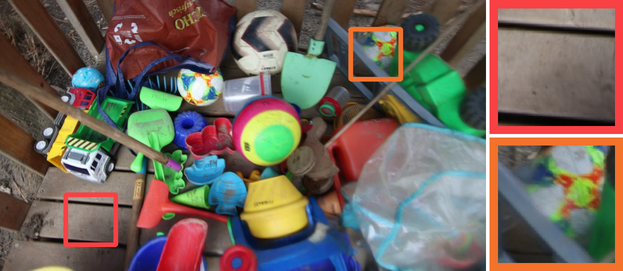} } & \raisebox{-0.5\height}{ \includegraphics[width=0.3167\linewidth]{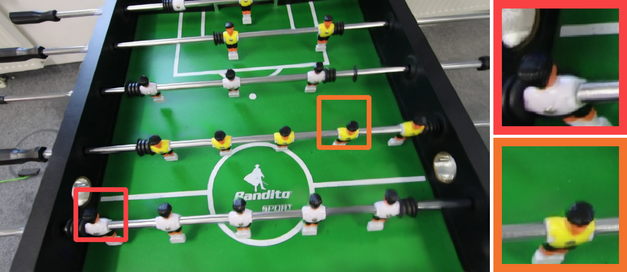} }\vspace{0.25em}\\
\rotatebox[origin=c]{90}{FVS \cite{Riegler2020Free}} & \raisebox{-0.5\height}{ \includegraphics[width=0.3167\linewidth]{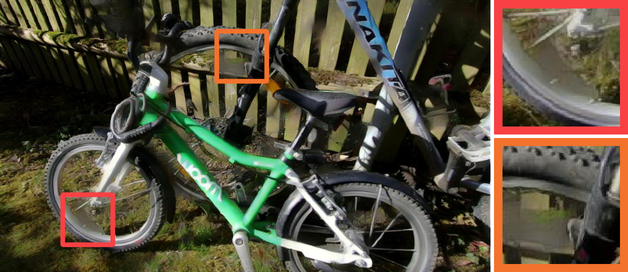} } & \raisebox{-0.5\height}{ \includegraphics[width=0.3167\linewidth]{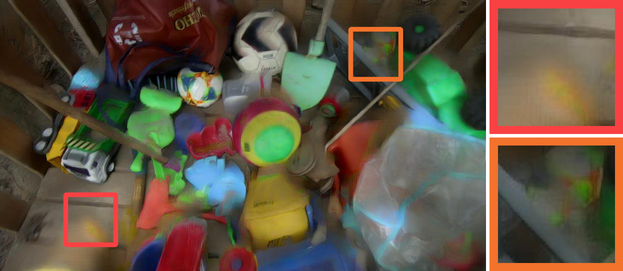} } & \raisebox{-0.5\height}{ \includegraphics[width=0.3167\linewidth]{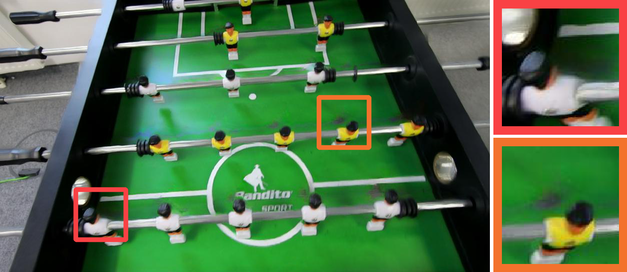} }\vspace{0.25em}\\
\rotatebox[origin=c]{90}{NeRF++ \cite{Zhang2020Nerfpp}} & \raisebox{-0.5\height}{ \includegraphics[width=0.3167\linewidth]{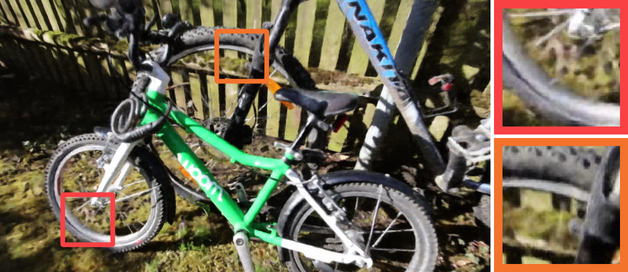} } & \raisebox{-0.5\height}{ \includegraphics[width=0.3167\linewidth]{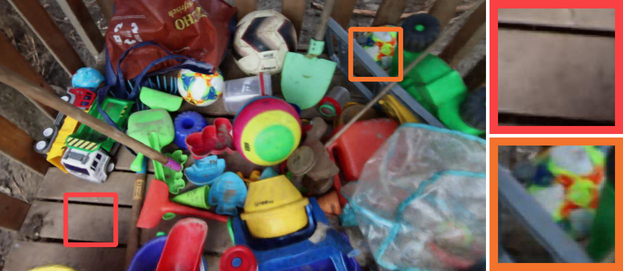} } & \raisebox{-0.5\height}{ \includegraphics[width=0.3167\linewidth]{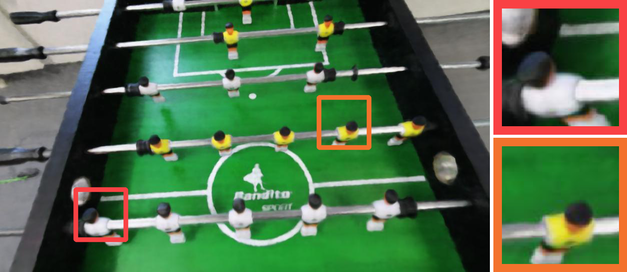} }\vspace{0.25em}\\
\rotatebox[origin=c]{90}{NPBG \cite{Aliev2020Neural}} & \raisebox{-0.5\height}{ \includegraphics[width=0.3167\linewidth]{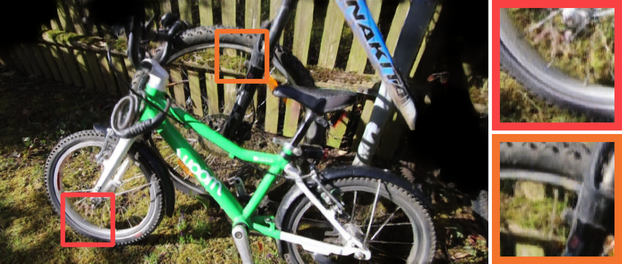} } & \raisebox{-0.5\height}{ \includegraphics[width=0.3167\linewidth]{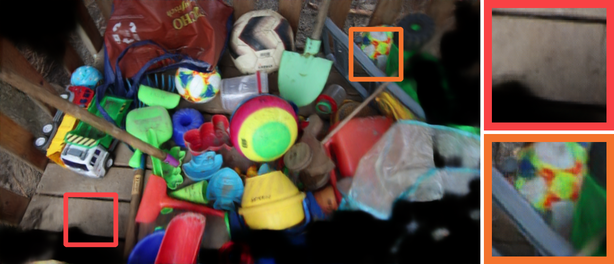} } & \raisebox{-0.5\height}{ \includegraphics[width=0.3167\linewidth]{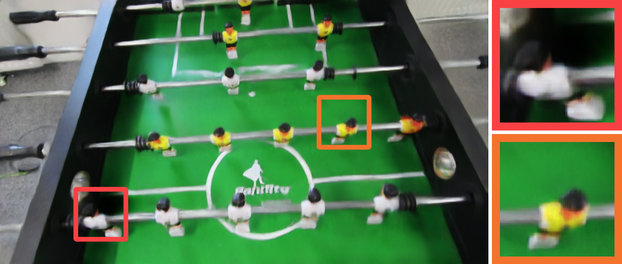} }\vspace{0.25em}\\
 & Bike & Sandbox & Soccertable\\
\end{tabular}
\endgroup\caption{
        \textbf{Qualitative results on the FVS dataset.}
        Comparison of SVS to the best-performing prior methods.
    }
    \label{fig:fvs}
\end{figure*}

We show qualitative results for the FVS dataset~\cite{Riegler2020Free} in Figure~\ref{fig:fvs}.
We observe that our method yields higher-fidelity results.
As this dataset contains some frames that exhibit some motion blur, we noticed that our results are at times sharper than the ground truth.

\begin{figure*}
    \center
    \begingroup
\setlength{\tabcolsep}{-0.1em}
\renewcommand{\arraystretch}{1}
\begin{tabular}{c c c c c c}
\rotatebox[origin=c]{90}{GT} & \raisebox{-0.5\height}{ \includegraphics[width=0.3167\linewidth]{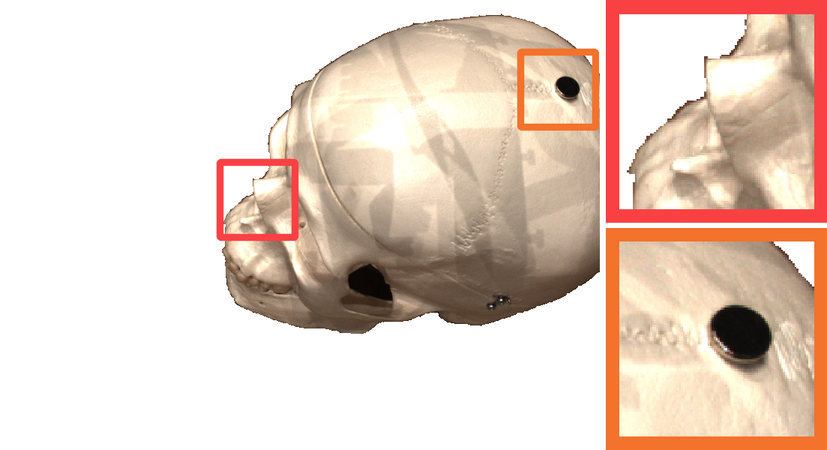} } & \raisebox{-0.5\height}{ \includegraphics[width=0.3167\linewidth]{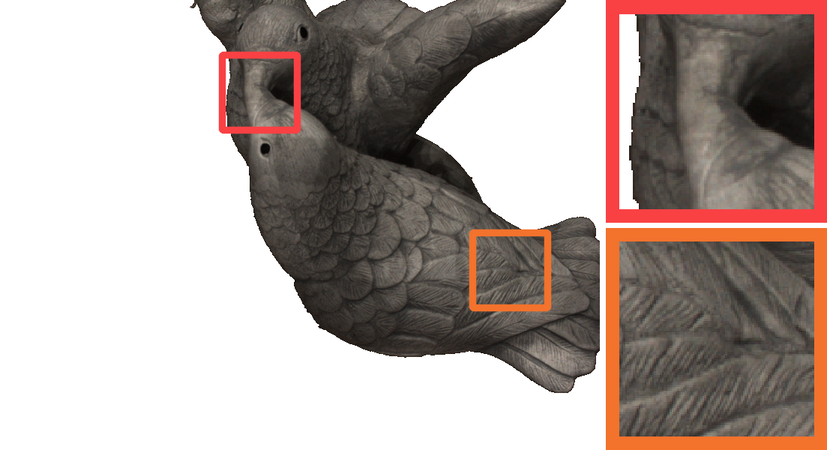} } & \raisebox{-0.5\height}{ \includegraphics[width=0.3167\linewidth]{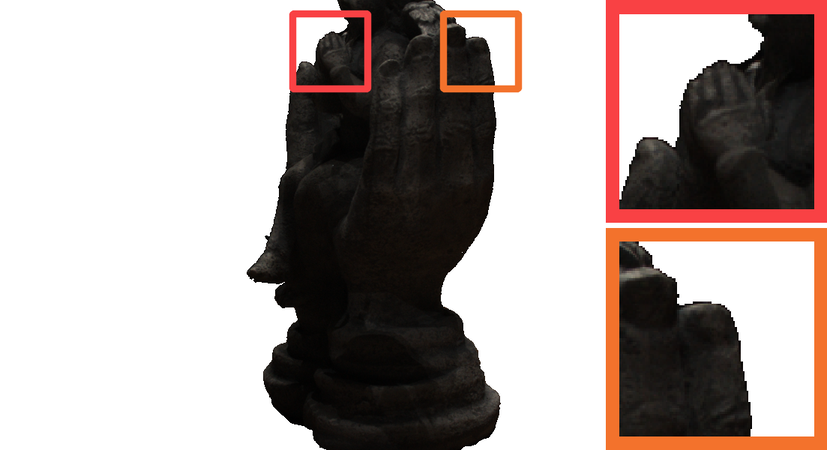} }\vspace{0.25em}\\
\rotatebox[origin=c]{90}{Ours} & \raisebox{-0.5\height}{ \includegraphics[width=0.3167\linewidth]{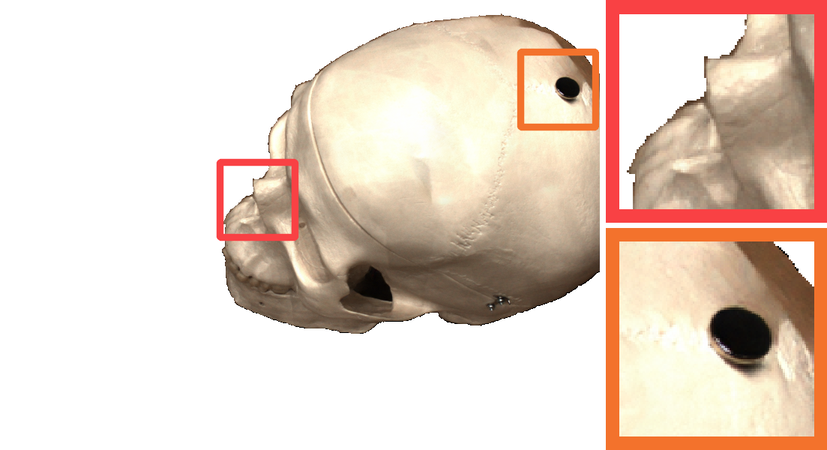} } & \raisebox{-0.5\height}{ \includegraphics[width=0.3167\linewidth]{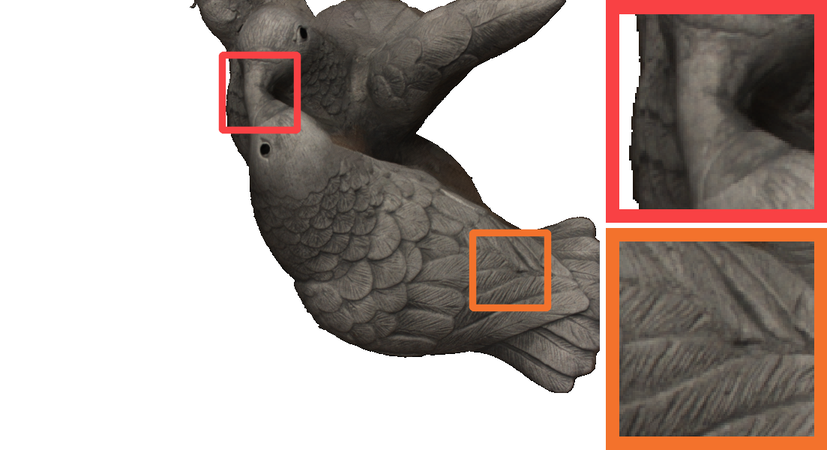} } & \raisebox{-0.5\height}{ \includegraphics[width=0.3167\linewidth]{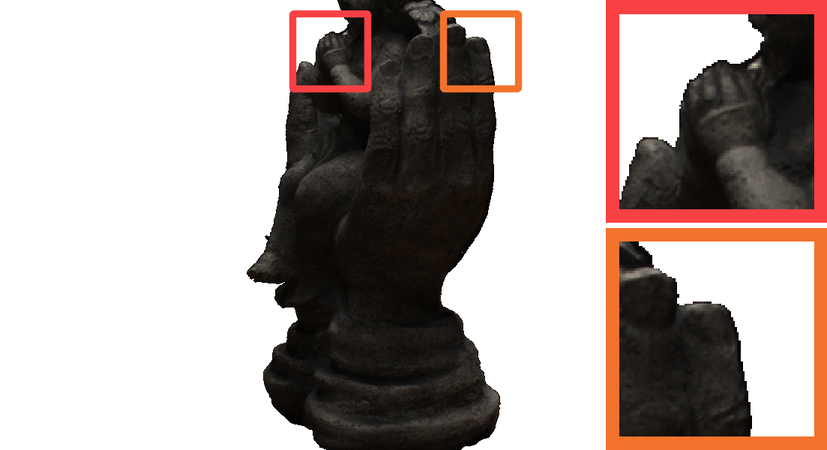} }\vspace{0.25em}\\
\rotatebox[origin=c]{90}{FVS \cite{Riegler2020Free}} & \raisebox{-0.5\height}{ \includegraphics[width=0.3167\linewidth]{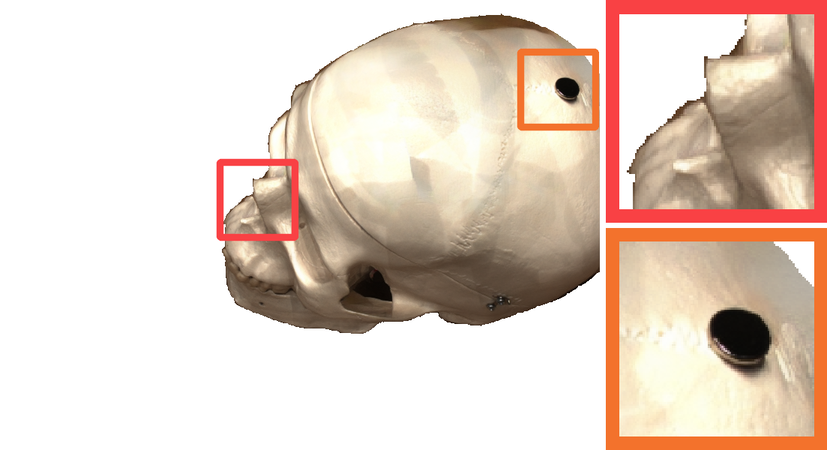} } & \raisebox{-0.5\height}{ \includegraphics[width=0.3167\linewidth]{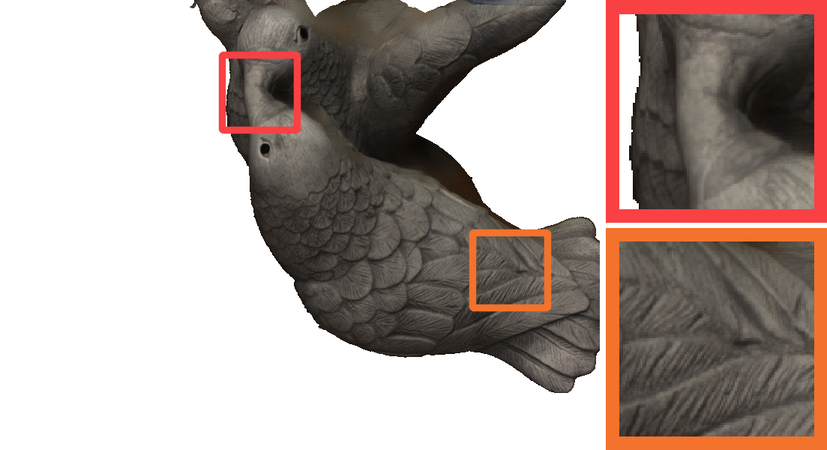} } & \raisebox{-0.5\height}{ \includegraphics[width=0.3167\linewidth]{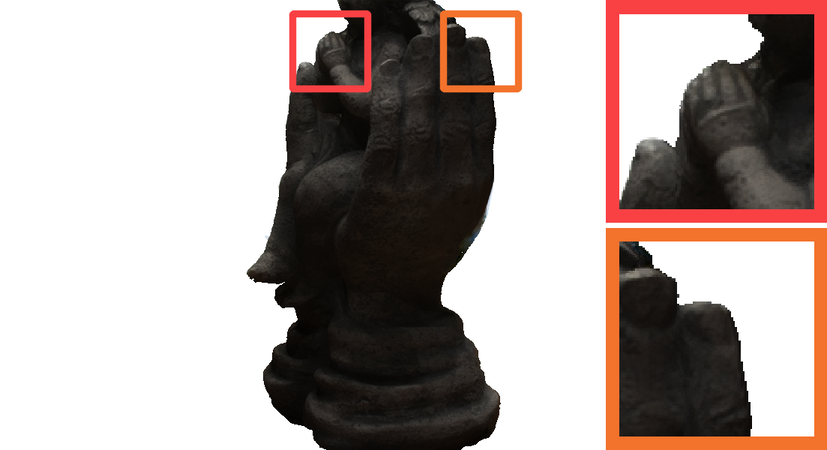} }\vspace{0.25em}\\
\rotatebox[origin=c]{90}{NeRF \cite{Mildenhall2020Nerf}} & \raisebox{-0.5\height}{ \includegraphics[width=0.3167\linewidth]{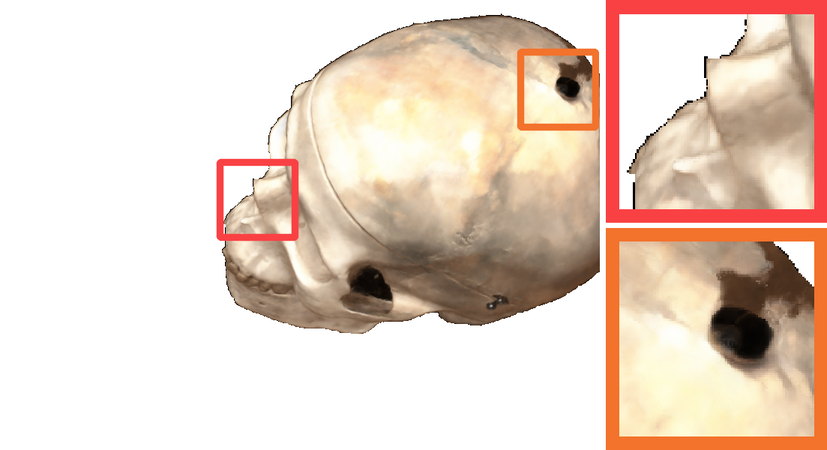} } & \raisebox{-0.5\height}{ \includegraphics[width=0.3167\linewidth]{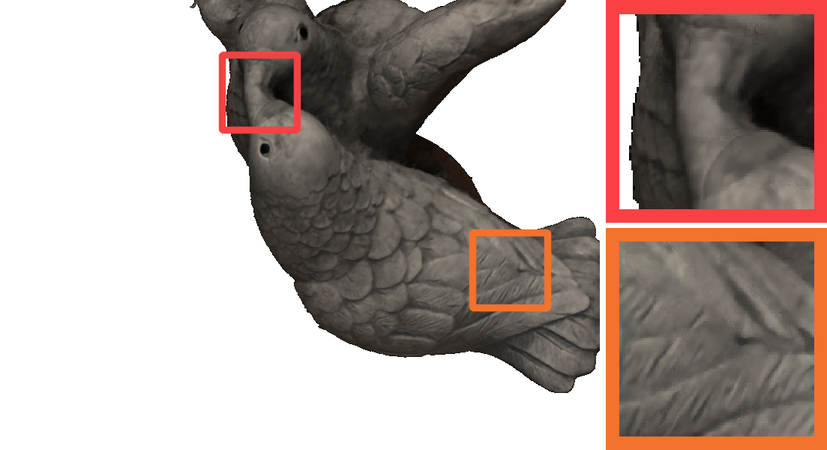} } & \raisebox{-0.5\height}{ \includegraphics[width=0.3167\linewidth]{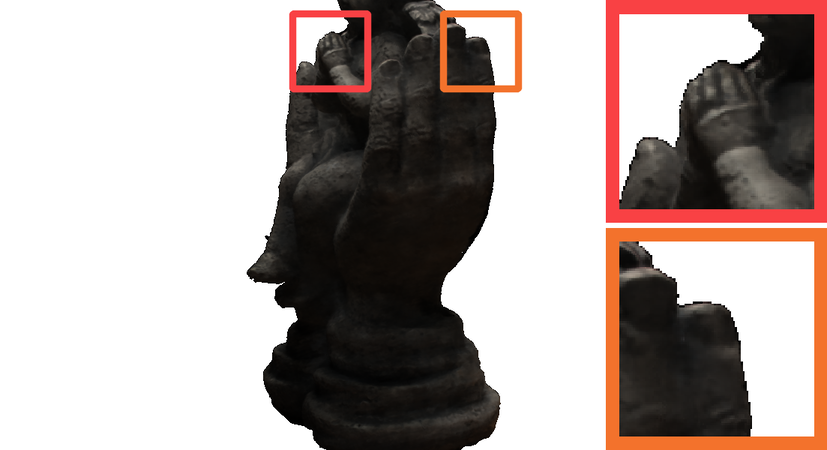} }\vspace{0.25em}\\
\rotatebox[origin=c]{90}{NPBG \cite{Aliev2020Neural}} & \raisebox{-0.5\height}{ \includegraphics[width=0.3167\linewidth]{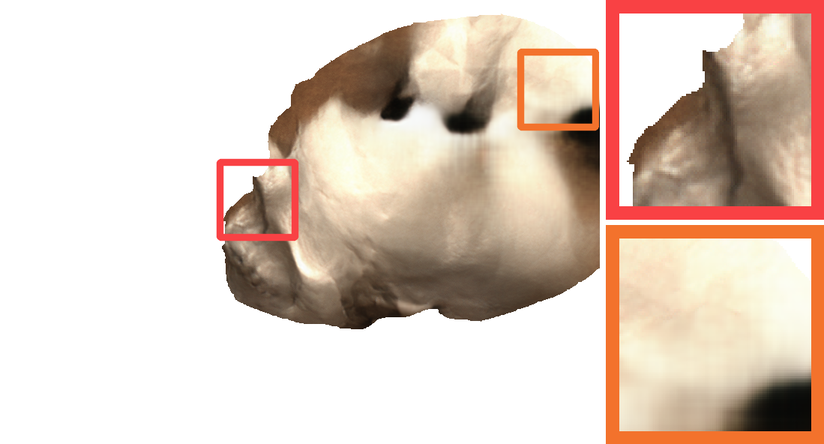} } & \raisebox{-0.5\height}{ \includegraphics[width=0.3167\linewidth]{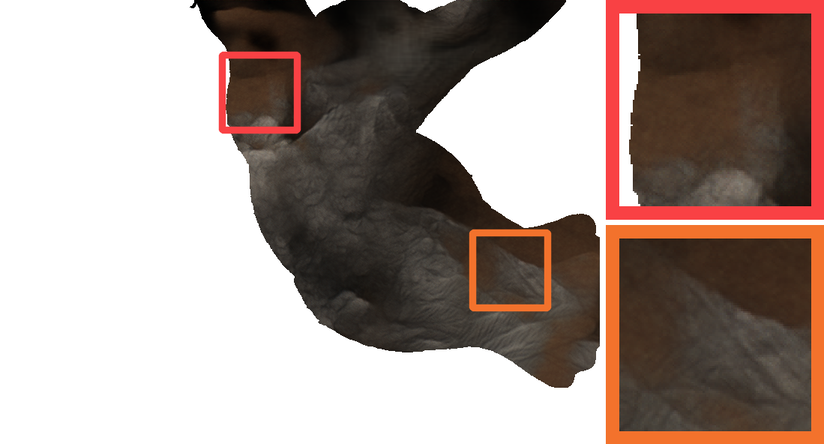} } & \raisebox{-0.5\height}{ \includegraphics[width=0.3167\linewidth]{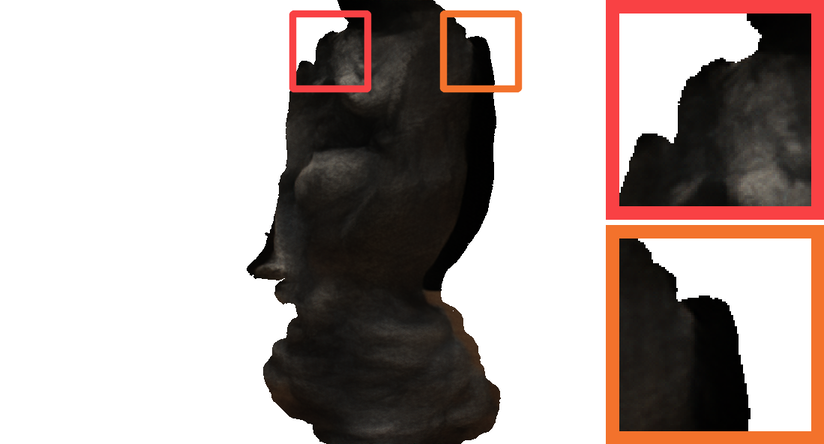} }\vspace{0.25em}\\
 & Scene 65 & Scene 106 & Scene 118\\
\end{tabular}
\endgroup\caption{
        \textbf{Qualitative results on DTU.}
        Comparison of SVS to the best-performing prior methods.
    }
    \label{fig:dtu}
\end{figure*}

Figure~\ref{fig:dtu} shows qualitative extrapolation results for the DTU dataset~\cite{Aanaes2016DTU}.
Note that the ground-truth images have artifacts due to shadows from the camera setup (e.g., top of the skull). These artifacts are not visible in our synthesized images, which sometimes look better than the ground truth for this reason.

Please see the supplementary video for sequences.

\mysuppsection{Runtimes}
In this section, we list the runtimes of our method and a selection of state-of-the-art methods.
The numbers below are for a typical scene from the Tanks and Temples dataset~\cite{Knapitsch2017Tanks}.

We start with a breakdown of our method.
We erect the geometric scaffold using COLMAP~\cite{Schoenberger2016SfM,Schoenberger2016Pixelwise}.
Structure-from-motion takes $\myunaryless 8$~minutes (including feature extraction, feature matching, triangulation, and bundle adjustment), multi-view stereo takes $\myunaryless 43$~minutes, pointcloud fusion takes $\myunaryless 14$~minutes, and Delaunay-based surface reconstruction takes $\myunaryless 32$~minutes.
This adds up to $\myunaryless 97$~minutes for erecting the geometric scaffold.
We also encode all source images, which takes $\myunaryless 25$~seconds.
Given a novel viewpoint, our method takes $\myunaryless 1$~second to synthesize an image.
This can be sped up further, as our current implementation loads the encoded images from RAM to GPU memory for each novel target view. If the encoded images are already in GPU memory, image synthesis takes $\myunaryless 0.2$~seconds.

NPBG~\cite{Aliev2020Neural} is based on a reconstructed point cloud of the scene.
As outlined above, this can be computed in $\myunaryless 65$~minutes.
Then, the NPBG representation has to be fitted to the scene.
Starting from a pretrained rendering network, training for $10$ epochs takes in total $\myunaryless 31$~minutes.
As all feature vectors are kept in GPU memory, synthesizing novel views is fast, taking $\myunaryless 0.1$~seconds on average.

NeRF++~\cite{Zhang2020Nerfpp} requires less geometric information, only the camera poses and the sparse point cloud from structure-from-motion.
As shown above, this can be computed in $\myunaryless 8$~minutes.
Then, NeRF++ has to be fitted to the given scene.
Optimizing it for 50,000 iterations takes $\myunaryless 24$~hours.
To synthesize a novel target image from NeRF++ requires $\myunaryless 71$~seconds.

FVS~\cite{Riegler2020Free} is based on the same geometric scaffold as our method, which can be erected in $\myunaryless 97$~minutes.
Mapping $7$ source images per novel target view and blending them via the recurrent network takes on average $\myunaryless 0.5$~seconds.